\documentclass[10pt,twocolumn,letterpaper]{article}

\usepackage[]{wacv} 

\usepackage{graphicx}
\usepackage{amsmath}
\usepackage{amssymb}
\usepackage{booktabs}
\usepackage{mathtools}

\usepackage[accsupp]{axessibility} 

\usepackage[pagebackref,breaklinks,colorlinks]{hyperref}

\usepackage{amsfonts}
\usepackage{adjustbox}
\usepackage{bm} 
\usepackage{makecell}
\usepackage{multirow}

\usepackage{pifont}
\usepackage[capitalize]{cleveref}
\crefname{section}{Sec.}{Secs.}
\Crefname{section}{Section}{Sections}
\Crefname{table}{Table}{Tables}
\crefname{table}{Tab.}{Tabs.}

\def\wacvPaperID{227} 
\def\confName{WACV}
\def\confYear{2025}

\def\etal{\emph{et al}.}

\newcommand{\diff}[1]{{#1}\xspace}

\DeclareMathOperator*{\argmin}{argmin}

\newcommand{\doubleunderline}[1]{\underline{\underline{#1}}}

\newcommand{\nickname}[0]{FaVoR\xspace}
\newcommand{\pnpransac}[0]{Render+PnP-RANSAC\xspace}

\usepackage{xcolor}
\definecolor{somegray}{rgb}{0.5, 0.5, 0.5}
\newcommand{\darkgrayed}[1]{\textcolor{somegray}{#1}}
\makeatletter
\newcommand*\titleheader[1]{\gdef\@titleheader{#1}}
\AtBeginDocument{%
  \let\st@red@title\@title
  \def\@title{%
    \vskip-3.5em
    \bgroup\normalfont\large\centering\@titleheader\par\egroup
    \vskip1.5em\st@red@title}
}
\makeatother

\titleheader{\darkgrayed{This paper has been accepted for publication at the IEEE/CVF Winter Conference on Applications of Computer Vision (WACV), Tucson, US, 2025. \copyright IEEE}}

\title{FaVoR: Features via Voxel Rendering for Camera Relocalization}

\author{Vincenzo Polizzi$^{1}$ \quad Marco Cannici$^{2}$ \quad Davide Scaramuzza$^{2}$ \quad Jonathan Kelly$^{1}$ \\[1mm]
$^{1}$ University of Toronto, $^{2}$ University of Zurich
\\[1mm]
{\tt\small \{vincenzo.polizzi,jonathan.kelly\}@robotics.utias.utoronto.ca},
{\tt\small cannici@ifi.uzh.ch}
}

\begin{document}

\maketitle
\begin{abstract}
Camera relocalization methods range from dense image alignment to direct camera pose regression from a query image. %
Among these, sparse feature matching stands out as an efficient, versatile, and generally lightweight approach with numerous applications.
However, feature-based methods often struggle with significant viewpoint and appearance changes, leading to matching failures and inaccurate pose estimates.
To overcome these limitations, we propose a novel approach that leverages a globally sparse but locally dense 3D representation of 2D features.
By tracking and triangulating landmarks over a sequence of frames, we construct a sparse voxel map optimized to render image patch descriptors observed during tracking.
Given an initial pose estimate, we first synthesize descriptors from the voxels using volumetric rendering and then perform feature matching to estimate the camera pose.
This method enables the generation of descriptors for unseen views, enhancing robustness to viewpoint changes.
We evaluate our method on the 7-Scenes and Cambridge Landmarks datasets.
Our results show that our approach significantly outperforms existing state-of-the-art feature representation techniques %
in indoor environments, achieving up to a 39\% improvement in median translation error. 
Additionally, our approach yields comparable results to other methods for outdoor scenes but with lower computational and memory footprints.

\end{abstract}



\noindent\textbf{Supplementary Material:} For additional details, please visit: \url{https://papers.starslab.ca/favor}

\section{Introduction}
\begin{figure*}[t]
    \includegraphics[width=\textwidth, trim=0 0 0 0.8in, clip]{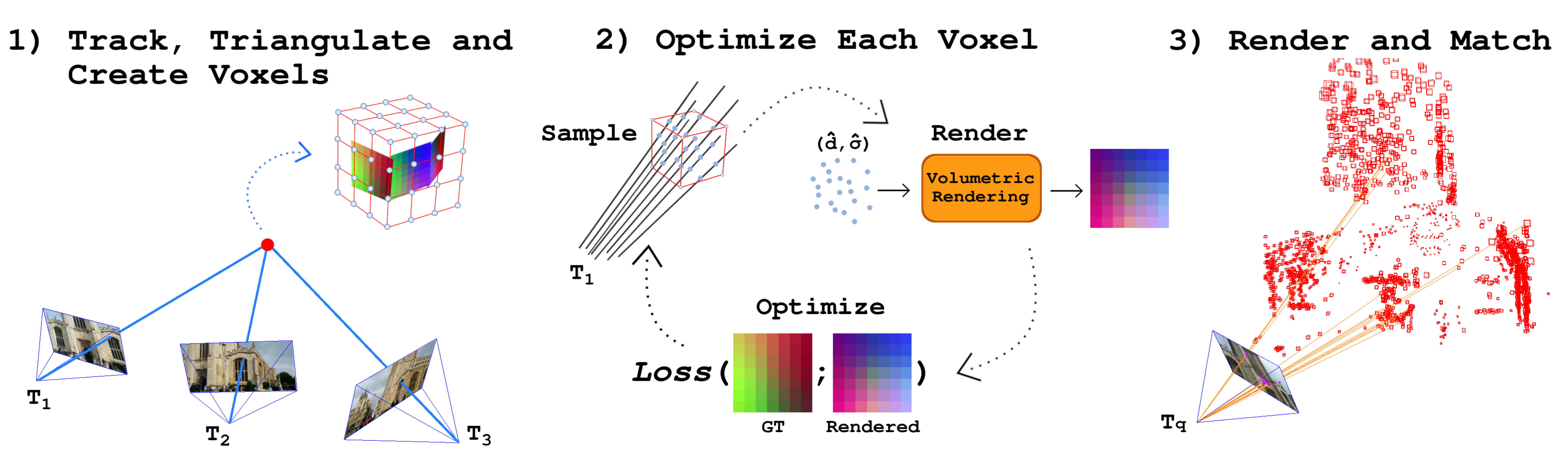} \hfill
    \vspace{-3ex}
    \caption{
        \textbf{Schematic representation of the proposed method.} 1) We track and triangulate feature points to create a voxel representation for persistent landmarks, i.e. landmarks observed by a large number of views. After that, 2) the voxels are optimized to render the descriptors patches extracted during the feature tracking. At test time, 3) the voxels are queried to render the descriptors as seen from a given query pose, and the 2D-3D matches between the query image and the landmarks are found and used to perform pose estimation.
    }
    \label{fig:eyecatcher}
\end{figure*}

\label{sec:intro}

Despite extensive research, visual localization remains a significant challenge, particularly under wide viewpoint and appearance changes. Various strategies have been proposed to improve localization performance in difficult scenarios. Sequence-based place recognition, for instance, exploits consistency across frames to identify previously visited locations~\cite{Milford12icra}, while geometric constraints derived from point cloud alignment and 3D-3D matching can provide additional, useful information~\cite{Maffra17ic3d, Maffra19ral, Yang21TRO}. Ultimately, the effectiveness of many of these strategies depends upon the reliability and robustness of low-level feature matching, which is used extensively to determine the camera pose. %
Recent efforts to enhance visual localization performance have explored techniques based on neural radiance fields (NeRF)~\cite{mildenhall2021nerf}.
These approaches build upon dense or semi-dense simultaneous localization and mapping (SLAM) frameworks~\cite{mildenhall2021nerf, Sucar2021ICCV, rosinol2023nerf, chung2022orbeez, zhu2024nicer} by either synthesizing frames for feature point extraction or by directly performing photometric alignment.
Other techniques exploit neural view synthesis to render a dense descriptor space in order to localize in a known map~\cite{moreau2023crossfire, liu2023nerfloc}, or to capture the appearance variations of a sparse set of feature points observed from a query viewpoint~\cite{germain2022cvpr}.
While these methods do improve performance, they come with some significant drawbacks.
The dense descriptor representation~\cite{liu2023nerfloc, moreau2023crossfire} often requires more training time and memory, since a dense and fully optimized radiance field must be learned prior to deployment. 
Furthermore, works that synthesize sparse descriptors~\cite{germain2022cvpr} struggle to render descriptors with a large number of channels.

Building on these insights, we introduce \nickname, a new feature rendering method that exploits a pre-trained neural network to extract %
features and a sparse voxel-based formulation to encode and render feature descriptors in 3D space. Our scene representation, illustrated in~\Cref{fig:eyecatcher}, is globally sparse but locally dense, allowing for efficient extraction of view-conditioned 3D point descriptors from any query camera pose. 
During training, we extract and track keypoints across a sequence of frames, then triangulate 3D landmarks using known camera poses. This approach mirrors existing online localization pipelines~\cite{Mourikis07icra}, which are inherently sequential.
Each landmark is represented as a voxel and optimized through volumetric rendering, allowing the associated descriptor to be rendered from novel views. In turn, we are able to carry out low-level feature matching under wide viewpoint changes. Unlike prior techniques that learn dense radiance fields~\cite{moreau2023crossfire, liu2023nerfloc}, \nickname encodes sparse landmark descriptors only, offering a favorable tradeoff between descriptor robustness and resource efficiency.
\nickname reduces the computational burden associated with more dense models, improving scalability in practical applications. 
Moreover, instead of using a neural network to render descriptors~\cite{moreau2023crossfire, liu2023nerfloc, germain2022cvpr}, we employ an explicit voxel representation and trilinear interpolation for rendering, accelerating both the training and rendering processes. %
We extensively evaluate our approach on the 7-Scenes~\cite{shotton13cvpr} and Cambridge Landmarks~\cite{Kendall15iccv} datasets. 
\nickname significantly outperforms state-of-the-art implicit feature rendering approaches on the indoor 7-Scenes dataset \cite{shotton13cvpr}, achieving up to a 39\% reduction in median translation error, and delivers comparable performance on outdoor scenes.

To summarize, we make the following contributions.
\begin{itemize}
\itemsep=-2pt
    \item We present \nickname, a sparse, voxel-based algorithm to render feature points tracked over a sequence of frames, that does not require learning a dense volumetric representation of the scene. %
    \item  We show how to render high-dimensional descriptors, which provide superior view-invariance, and test our descriptor representation using different state-of-the-art feature points extractors with different numbers of channels: 64, 96, 128~\cite{Zhao2022ALIKE}, and 256~\cite{detone2018superpoint}.%
    \item We demonstrate using the 7-Scenes~\cite{shotton13cvpr} and Cambridge Landmarks~\cite{Kendall15iccv} datasets that our system is capable of rendering descriptors faithfully from unseen views; \nickname improves camera pose estimation performance compared to other state-of-the-art implicit descriptors rendering techniques~\cite{germain2022cvpr, moreau2023crossfire, liu2023nerfloc}.
    \item We release our codebase and pre-trained models to the community.
\end{itemize}

\section{Related Work}
\label{sec:related}
We divide traditional visual localization systems into three categories according to the scene representation employed: image-based, structure-based, and hybrid. This section provides an overview of related work in these categories and contrasts traditional methods with newer techniques (collectively called \emph{scene feature renderers}) that synthesize new descriptors based on a query pose.

\textbf{Image-based} methods (IBMs) for visual localization use appearance information only to determine the camera pose for a given query image.
A set of known images and associated camera poses is required, which may simply be stored in a database and searched when the query image is fed to the system, or used to train a neural network that learns to regress camera pose.
Hence, this category includes absolute~\cite{Kendall15iccv, naseer2017deep, shavit2021learning, chen20213DV} and relative~\cite{balntas2018relocnet, laskar2017camera, turkoglu2021visual} pose regression, pose triangulation, and pose interpolation~\cite{torii11iccvw} approaches.
Most recently, neural radiance fields (NeRF)~\cite{mildenhall2021nerf} have been employed to enhance the training of image-based models by providing synthetic views~\cite{moreau2022PMLR}.
Prior research has shown that these methods yield limited pose estimation accuracy ~\cite{Sattler19cvpr, zhou2020learn}, especially in large environments, due to the approximate nature of pose interpolation.
 
\textbf{Structure-based} methods (SBMs), on the other hand, use the 3D structure of the scene, typically computed via structure-from-motion (SfM), to recover camera pose through 2D-3D feature point matching.
These paradigms can deliver accurate 3D camera pose estimates, particularly when coupled with RANSAC~\cite{Fischler81cacm} or similar techniques for outlier rejection.
The 3D structure of the environment can also be learned end-to-end using neural networks, as demonstrated by SCoRe Forest~\cite{shotton13cvpr} and DSAC*~\cite{brachmann2021dsacstar}.
Generally, these methods are relatively slow, but recently, ACE~\cite{Alzugaray18threedv} has demonstrated fast convergence and better camera pose estimation results than its predecessors.
Our work exploits the sparsity of the structure-based representation but uses sequences of frames to track and triangulate landmarks, instead of relying on complete SfM models.

\textbf{Hybrid} methods (HMs) attempt to combine the advantages of the image and structure-based approaches.
For each image in a database, the set of 3D landmarks visible from the associated camera pose can be determined, and matching can be performed between a small set of candidate features and those extracted from the query image.
To retrieve the most similar image, full-image descriptors~\cite{torii2015densevlad} or a vocabulary of visual features~\cite{qader19BoW, Zhang18SPIE} can be used. %
\diff{Hybrid methods require significant memory to store the full SfM representation of the scene as well as the landmark descriptors, but they typically achieve the lowest relocalization error in large-scale environments~\cite{sarlin2019coarse}.
Some methods aim to reduce memory usage, but at the cost of increased relocalization error~\cite{scenesqueezer2022cvpr}.}
We follow a similar hybrid approach, rendering descriptors only for those landmarks that are likely to be visible from a given camera pose.

\textbf{Scene feature rendering} methods (SFRMs) form a newer class of visual localization techniques that directly render 3D point descriptors from novel views in order to localize.
These methods potentially overcome one of the major limitations of traditional feature extractors, that is, the lack of viewpoint invariance.
Although learning-based approaches have led to improvements in feature extraction and matching, the network training process still relies on 2D images, and hence the learned descriptors are closely tied to image appearance and viewpoint.
To address the problem of viewpoint invariance, FQN~\cite{germain2022cvpr} learns to model the appearance of feature points observed from different views, but struggles to render high-dimensional descriptors (see \cite[Section 6]{germain2022cvpr}).
Moreau et al.~propose CROSSFIRE in \cite{moreau2023crossfire}, a NeRF model that builds a dense scene representation in descriptor space, rather than in the RGB space, allowing for continuous descriptor rendering across different views.
Most recently, Liu et al.~describe NeRF-loc~\cite{liu2023nerfloc}, an approach that leverages the depth information provided by an RGB-D camera to create an accurate NeRF model and then to learn 3D descriptors that can be matched with new views via a multi-headed attention mechanism.
Despite the pose estimation accuracy obtained, these methods are limited by training time and memory usage, as they require a full 3D model of the environment. 
Consequently, they are unsuitable for large-scale and real-time localization tasks.
Our approach relies on volumetric rendering but for a sparse set of voxels only, derived from features tracked over a sequence of RGB input images.

\section{Methodology}
\label{sec:method}
We track a sparse set of landmarks through a sequence of training images.
We then use volumetric rendering to optimize a local, dense voxel grid representation of each landmark, rendering the 2D descriptor patches observed during tracking (see~\Cref{fig:eyecatcher}). %
\diff{At inference time, we iteratively refine the camera estimate by querying the optimized voxel set to determine each landmark's appearance (i.e., feature) based on an initial camera pose guess,  as proposed by~\cite{germain2022cvpr}.
The rendered descriptors are then matched with the features in the query image, and these correspondences are used to compute the camera pose by solving the PnP problem~\cite{kukelova2016gp3p} within a RANSAC scheme~\cite{Fischler81cacm, chum2003loransac}. 
We refer to this iterative process as \pnpransac}.
In the following subsections, we describe each step in more detail.
\subsection{Landmark Tracking}
\label{sec:tracking}

Consider a sequence of $M$ RGB images, each $H \times W$ pixels in size, $\mathbf{I}_{1,\hdots, M} \in \mathbb{R}^{H \times W \times 3}$, and a landmark $\pmb{\ell}_j \in \mathbb{R}^3$ in the world frame. Define $\mathcal{S}_j$ as the set of indices of training images that include $\pmb{\ell}_j$ (i.e., that observe $\pmb{\ell}_j$).
For each image $\mathbf{I}_i$, with $i \in \mathcal{S}_j$, there exists a keypoint $\mathbf{k}_{ij} \in \mathbb{R}^2$ that is the projection of $\pmb{\ell}_j$ onto $\mathbf{I}_i$.
The camera pose at $\mathbf{I}_i$ is $\mathbf{T}_i \in \mathrm{SE}(3)$ in the world frame.
We denote by $\mathcal{F}$ a feature extractor that, given image $\mathbf{I}_i$, provides keypoints and a dense descriptor map %
$\mathbf{D}_i \in \mathbb{R}^{H \times W \times C}$, where $C$ is the number of descriptor channels.
From $\mathbf{D}_i$, we crop patches of size $S \times S$ pixels, $\mathbf{P}_{ij} \in \mathbb{R}^{S \times S \times C}$, centered at each extracted keypoint $\mathbf{k}_{ij}$,
\begin{align}
    \mathbf{P}_{ij} = \text{crop}(\mathbf{D}_i, \mathbf{k}_{ij}, S).
\end{align}
For each landmark $\pmb{\ell}_j$, we store a \emph{track} that includes the camera poses $\mathbf{T}_i$ for the sequences of images that include $\pmb{\ell}_j$, the set of keypoints $\mathbf{k}_{ij}$, and the corresponding descriptor patches $\mathbf{P}_{ij}$.

Given a track, we triangulate the landmark position $\pmb{\ell}_j$ in the world frame.
An initial estimate of $\pmb{\ell}_j$ is found using the direct linear transform algorithm~\cite{Hartley2004} and refined by minimizing the reprojection error in a Levenberg–Marquardt optimization technique. %
To account for the presence of outliers, we apply a robust cost in the optimization process~\cite{geman1986bayesian}.
To ensure the numerical stability of the optimization, we used the inverse depth parameterization of the landmark position in the camera frame.
The measurement model and cost function are described in detail in our supplementary materials. 
\diff{Our approach is designed to integrate seamlessly into standard front-end pipelines for robot localization, making it suitable for existing online vision-based localization systems like visual-inertial odometry (VIO)~\cite{Mourikis07icra, Delaune20xvio}.}
\subsection{Voxel Creation}
We instantiate a new voxel $\mathcal{V}_j$, centred at $\pmb{\ell}_j$, for each track that is longer than a certain, minimum length\diff{, as common localization system do for persistent landmarks selection~\cite{Mourikis07icra}.}
The landmark associated with a track must be visible from many poses to provide useful localization information. 
Each voxel $\mathcal{V}_j$ consists of a grid, containing a number of smaller subvoxels, with a resolution of $R \times R \times R$.
Each node (i.e., vertex) of the resulting grid stores a vector of size $C$, corresponding to the descriptor channels provided by $\mathcal{F}$. %
To determine the appropriate size of the voxel (and its subvoxels), we first compute the Euclidean distance $l_{ij}$ from $\mathbf{T}_i$ to the point $\pmb{\ell}_j$, for each patch $\mathbf{P}_{ij}$.
Then, given the patch size of $S \times S$ pixels, we estimate the voxel size $s_{\mathcal{V}_j}$ (in metres) as
\vspace*{-3mm}
\begin{align}
    s_{\mathcal{V}_j} = \min_{i \in \mathcal{S}_j}(S \cdot \frac{l_{ij}}{f}),
\end{align}
where $f$ is the camera focal length.
We choose the minimum voxel size since this is sufficient to capture the information required to render the descriptor associated with the keypoint $\mathbf{k}_{ij}$.
Additionally, each voxel $\mathcal{V}_j$ is associated with a density grid of the same resolution and size as the descriptors grid, but with nodes of size one instead of $C$. %
The density grid is used in the volumetric rendering process described in~\Cref{sec:voxel_training}.

\subsection{Descriptor Learning and Rendering}
\label{sec:voxel_training}

Once the voxels are created, we can train our system to render the descriptor patches observed along the associated tracks.
The training process is similar to the method described in~\cite{mildenhall2021nerf}, but with notable differences, such as the absence of a multi-layer perceptron (MLP) for view-dependent rendering.
For all patches and poses $i \in \mathcal{S}_j$, we trace rays from the camera center $\mathbf{T}_i$ that pass through each element of the patch $\mathbf{P}_{ij}$.
Each of the rays $\mathbf{r}$ intersects the voxel grid $\mathcal{V}_j$ and its associated density grid at two points, one closer to the camera, $\mathbf{p}_n$, and one farther away, $\mathbf{p}_f$. We take $N$ samples, $\mathbf{d}_{t} \in \mathbb{R}^C$ 
and $\hat{\sigma_{t}} \in \mathbb{R}$, with $t = 1, \dots, N$, along the ray between the two intersection points, using trilinear interpolation from both $\mathcal{V}_j$ and the density grid, respectively (see~\Cref{fig:eyecatcher}).%
This process follows the volume rendering approach proposed in~\cite{max1995optical}, but instead of rendering an RGB color, we render a feature descriptor,
\begin{align}
    \delta &= \frac{||\mathbf{p}_f - \mathbf{p}_n ||_2}{N},\label{eq:vol_rendering_0}\\
    \hat{\mathbf{d}}_{ij}^{uv} &= \sum_{t=1}^{N} T_t \left(1-\exp{(-\hat{\sigma_{t}}\delta)}\right)\, \mathbf{d}_{t}\label{eq:vol_rendering_1}, \\ 
    T_t &= \prod_{l=1}^{t-1} \exp{(-\hat{\sigma_{l}}\delta)}),
    \label{eq:vol_rendering_2}
\end{align}
where $\hat{\mathbf{d}}_{ij}^{uv}$ is the estimated descriptor for landmark $\pmb{\ell}_j$ seen from the pose $\mathbf{T}_i$ at the pixel location $(u,v)$ in patch $\mathbf{P}_{ij}$.
To learn the descriptor and density grids, we want to ensure that $\hat{\mathbf{d}}_{ij}^{uv}$, in the descriptor vector space, is as close as possible in norm and direction to the ground truth descriptors $\mathbf{d}_{ij}^{uv} \in \mathbf{P}_{ij}$ extracted by $\mathcal{F}$ for all the $(u, v)\in \{(0,0), (0,1), \dots,(S, S)\}$ in the patch.
We use the mean squared error (MSE) and cosine similarity losses to drive the rendered descriptor's norm and direction as close as possible to the target feature, along with total variation regularization~\cite{rudin1994tv} to ensure a smooth representation of the patches $\mathbf{P}_{ij}$. For the density, we use the same entropy loss as described in~\cite{Sun22cvpr}.
In our supplementary material, we provide more details on the losses used.
Note that we apply the training process to each voxel independently of all the other voxels, enabling parallelization of the training process.

To render the descriptors from the voxels in the scene observed from a novel view, our system requires an initial guess $\mathbf{\hat{T}}_q$ for the query pose $\mathbf{T}_q$ that we aim to estimate.
We assume that this estimate is provided, \diff{as is typical in robotics localization systems where a prior pose estimate is available.}
Given the pose $\mathbf{\hat{T}}_q$ and the set of voxels $\mathcal{V} =\{\mathcal{V}_0, \mathcal{V}_1, \dots, \mathcal{V}_J\}$ in the scene, we can render the descriptors for all of the landmarks that are visible from a given query pose (this rendering may include points that are occluded because of the lack of depth information).
For each $\mathcal{V}_j$, we trace a ray from the query camera pose $\mathbf{\hat{T}}_q$ to the voxel-grid center $\pmb{\ell}_j$ (i.e., the position of the landmarks).
Following Equations (\ref{eq:vol_rendering_0}), (\ref{eq:vol_rendering_1}) and (\ref{eq:vol_rendering_2}), we then perform volumetric rendering along the ray to obtain the expected descriptor as seen from $\mathbf{\hat{T}}_q$.

\subsection{2D-3D Matching and Pose Estimation}
\label{sec:matching_pose}

\begin{figure*}[ht]
    \centering
    \begin{subfigure}[b]{0.33\linewidth}
        \includegraphics[width=\textwidth]{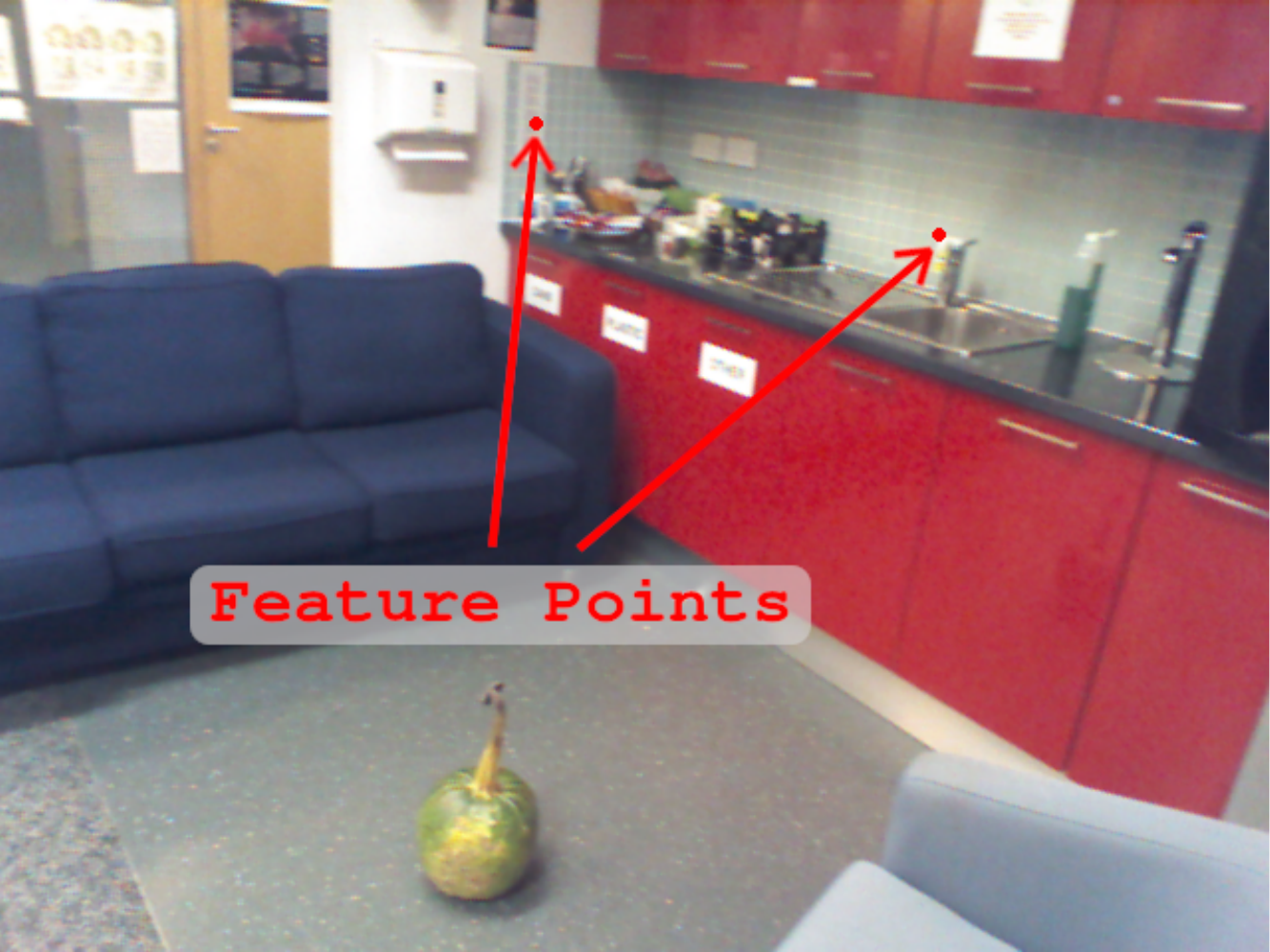} 
        \caption{}
        \label{fig:short-a}
    \end{subfigure}
    \begin{subfigure}[b]{0.33\linewidth}
        \includegraphics[width=\textwidth]{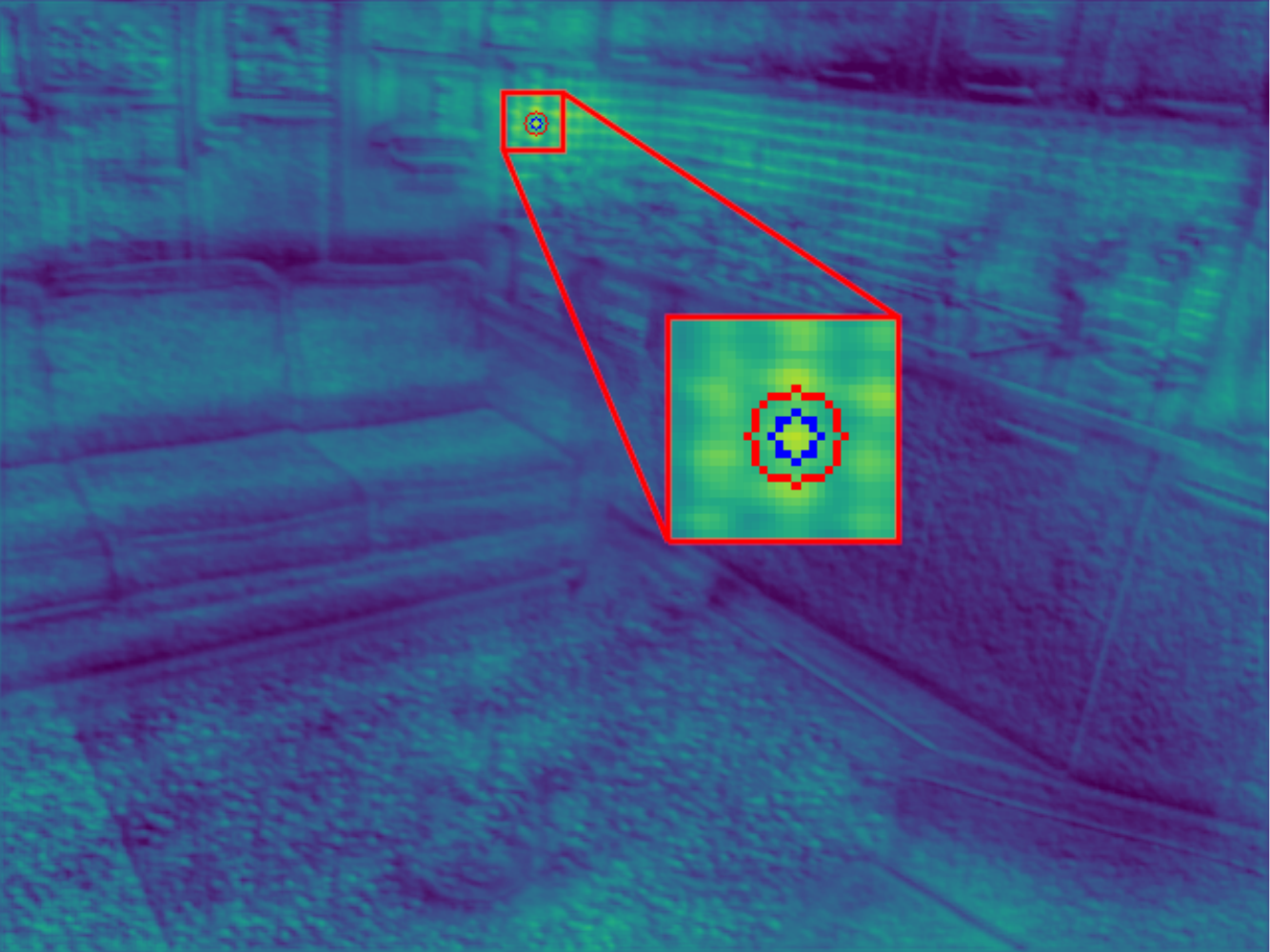} 
        \caption{}
        \label{fig:short-b}
    \end{subfigure}
    \begin{subfigure}[b]{0.33\linewidth}
        \includegraphics[width=\textwidth]{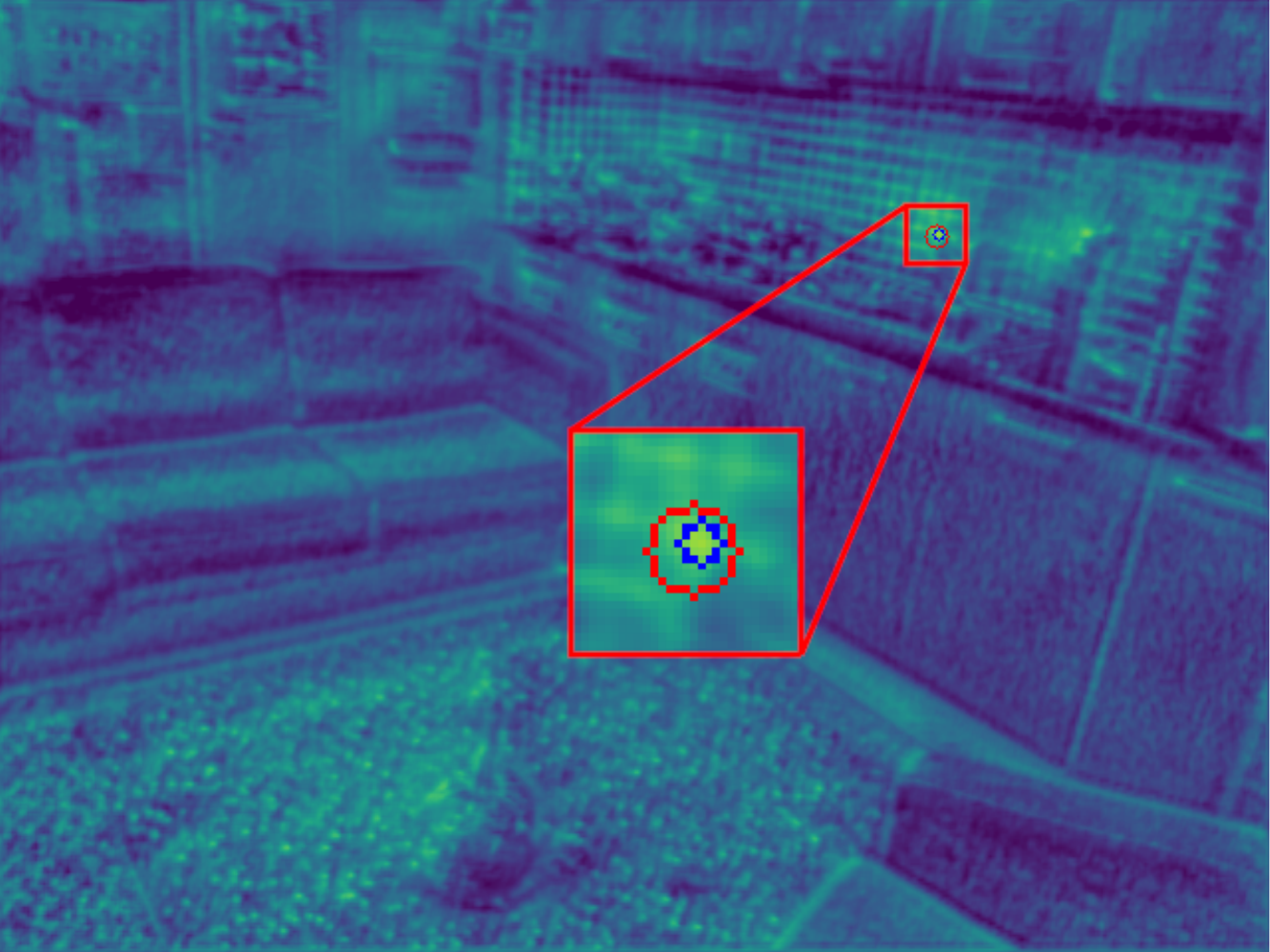} 
        \caption{}
        \label{fig:short-c}
    \end{subfigure}%
    \vspace*{-2mm}
    \caption{\label{fig:feature_map_matching}\textbf{Visualization of similarity response.}  We render a feature tracked during training using the Alike-l descriptor from an unseen view. On the left, a) displays the ground truth positions of the rendered feature points, obtained by projecting the triangulated landmarks on the camera plane, in red. At the same time, b) and c) show the similarity response between the rendered features and the target image dense descriptor map. The yellow colour indicates a strong response, concentrated around the feature positions shown in a), demonstrating the effectiveness of our descriptor rendering approach. The small circle in blue is the circle centre at the strongest response, the red circle is centred at the project landmark position. \diff{The top three response peaks per keypoint are 0.897, 0.868, and 0.836 for b) and 0.856, 0.786, and 0.737 for c) obtained by performing non-maxima suppression with a ray of three pixels, our method effectively renders descriptors that are distinctive in the image.}
    }
\end{figure*}

Once all the descriptors visible from $\mathbf{\hat{T}}_q$ are rendered, we can find 2D-3D correspondences with the query image $\mathbf{I}_q$.
A feature extractor generally finds sparse 2D keypoints in the query image and associates each of them with a descriptor extracted from a dense descriptor map given by $\mathcal{F}$~\cite{Zhao2023ALIKED,detone2018superpoint}.
To match our rendered descriptors to those extracted from the query image, we find the correspondences that provide the highest similarity score, above a certain threshold; that is, we compute a similarity matrix between the two descriptor sets, and after thresholding, we consider only the descriptor pairs with the maximum similarity response.
\Cref{fig:feature_map_matching} shows the similarity response of a rendered descriptor with the dense descriptor map extracted by the Alike-l~\cite{Zhao2022ALIKE} feature extractor from an unseen view.

Once all the rendered features are matched and the putative 2D-3D correspondences are available, we use PnP~\cite{kukelova2016gp3p} in a RANSAC~\cite{chum2003loransac} scheme to determine the camera pose.
Our feature representation allows us to render descriptors from novel views, \diff{enabling the usage of the iterative \pnpransac refinement procedure}.
This process highlights the distinctive characteristics of the proposed 3D feature representation. %
As the estimated query pose $\mathbf{\hat{T}}_q$ converges to $\mathbf{T}_q$, the rendered descriptors increasingly match the query image descriptors' appearance, resulting in a higher number of correspondences (see~\Cref{sec:experim}).

\section{Results}
\label{sec:experim}
\newcolumntype{C}[1]{>{\centering\arraybackslash}p{#1}}

\begin{table*}[t]
\begin{center}
    \begin{tabular}{l|l|C{1.5cm}|C{1.5cm}|C{1.5cm}|C{1cm}|C{1cm}|C{1cm}}
        \toprule %
        \multirow{2}{*}{
        \vspace{-1.2mm} Dataset} & 
        \multirow{2}{*}{
        \vspace{-2.3mm}
        \shortstack[l]{Feature\\ Extractor}}
         & \multicolumn{3}{c|}{Median Pose Error (cm, deg) $\downarrow$} & \multicolumn{3}{c}{Avg. Inliers per Image $\uparrow$} \\[0.6mm]
        \cline{3-8}
        & & 
        \vspace{-2.2mm}
        Iter 1 & 
        \vspace{-2.2mm}
        Iter 2 & 
        \vspace{-2.2mm}
        Iter 3 & 
        \vspace{-2.2mm}
        Iter 1 & 
        \vspace{-2.2mm}
        Iter 2 & 
        \vspace{-2.2mm}
        Iter 3 \\
        \midrule
        \multirow{2}{*}{7-Scenes}
        & \nickname\textsubscript{Alike-l} & 1.5, 0.4 & 1.4, 0.4 & \textbf{1.3, 0.4} & 49 & 54 & 54 \\
        & \nickname\textsubscript{SP} & 1.8, 0.5 & 1.6, 0.4 & \textbf{1.5, 0.4} & 73 & 81 & 81 \\
        \midrule
        \multirow{2}{*}{Cambridge}
        & \nickname\textsubscript{Alike-l} & 18.0, 0.3 & 16.8, 0.3 & \textbf{15.6, 0.3} & 205 & 222 & 221 \\
        & \nickname\textsubscript{SP} & 20.8, 0.4 & 19.6, 0.3 & \textbf{18.2, 0.3} & 200 & 218 & 218 \\
        \bottomrule
    \end{tabular}
    \end{center}%
    \vspace*{-3mm}
    \caption{\textbf{Comparison of different feature extractor networks}. We compare different feature extractors for pose estimation using \nickname rendering. Reported are the median translation and rotation error and the number of 2D-3D match inliers per image. The subcolumns indicate results obtained after 1 to 3 PnP-RANSAC scheme iterations. In \textbf{bold} are the best overall pose estimate results.
    }
    \label{tab:svfr_comparison_partial}
    \vspace{-2mm}
\end{table*}

We evaluate our system for camera relocalization in both indoor and outdoor environments using two well-established benchmarks:
the 7-Scenes~\cite{shotton13cvpr} and Cambridge Landmarks~\cite{Kendall15iccv} datasets.
We provide implementation details and information about our baseline comparisons in \Cref{subsec:details}.
In~\Cref{sec:evaluation}, we evaluate our method against existing state-of-the-art (SOTA) relocalization systems that use different scene representation paradigms.
We report on the memory footprint and discuss the computational requirements of \nickname compared to other systems in~\Cref{sec:memory_and_time_comparison}.
Lastly, in~\Cref{sec:rendering_capabilities}, we examine the rendering capabilities of \nickname as a function of the viewing angle.

\subsection{Implementation Details and Baselines}
\label{subsec:details}

Our implementation is based on PyTorch, building on the framework provided by~\cite{Sun22cvpr}.
We use Alike~\cite{Zhao2022ALIKE} and SuperPoint~\cite{detone2018superpoint} as feature extractors.
We test descriptors with 64 (Alike-t), 94 (Alike-s), 128 (Alike-n), 128 (Alike-l), and 256 (SuperPoint) channels.
Training is carried out for 2,000 epochs per voxel (landmark) using 1,024 rays per epoch, with the Adam optimizer~\cite{kingma2014adam} and a variable learning rate per subvoxel as in~\cite{Sun22cvpr}.
We examine the tradeoff between model size, rendering quality, expressed by the signal-to-noise ratio (PSNR), and grid resolution (details in the supplementary material), and chose a grid resolution of $3 \times 3 \times 3$ to balance memory usage and rendering performance.
Training a voxel on a $7\times7$ pixel descriptor patch takes about 10 seconds on an NVIDIA GeForce RTX 4060 laptop GPU\diff{, rendering a single descriptor takes about 1 ms.}
\diff{However, \nickname is not currently optimized for fast runtime operations. Since the voxels are independent of each other, the entire training and inference pipeline can be parallelized to significantly improve performance.}

We compare \nickname with similar SFR methods that render descriptors from unseen views: 
FQN~\cite{germain2022cvpr}, CROSSFIRE~\cite{moreau2023crossfire}, and NeRF-loc~\cite{liu2023nerfloc}, which closely align with our work in terms of scope, despite the different methodologies and requirements (see~\Cref{sec:related}).
FQN~\cite{germain2022cvpr} and CROSSFIRE~\cite{moreau2023crossfire} rely on large sets of points, using SfM and dense feature matching, respectively. NeRF-loc~\cite{liu2023nerfloc} matches each query image with 1,024 3D points. In contrast, our approach uses a smaller set of points. %
Additionally, while~\cite{liu2023nerfloc, moreau2023crossfire} require a fully trained neural rendering model to work, \nickname needs only a sequence of frames to train the sparse voxel representation of the landmarks.
Moreover, \nickname works with out-of-the-box feature extraction methods, without requiring a custom, scene-dependent network as in~\cite{moreau2023crossfire}.
\diff{By training \nickname using descriptors with different numbers of channels, we demonstrate that our method is scalable to different descriptor sizes, unlike~\cite{germain2022cvpr}.}
Also, while~\cite{germain2022cvpr} iterates the \diff{\pnpransac} multiple times, we use only three iterations, similarly to~\cite{moreau2023crossfire}. %
For comprehensiveness, we report the localization results of image-based~\cite{Kendall15iccv, brahmbhatt2018mapnet, shavit2022pae, moreau2022PMLR} and structure-based~\cite{wu2022scwls, brachmann2021dsacstar, brachmann2023ace} camera relocalization methods, as well as hybrid methods (see~\Cref{sec:related} for more details).
\diff{For the HM category, we report the results from HLoc~\cite{sarlin2019coarse}, based on SuperPoint~\cite{detone2018superpoint} (SP) and SuperGlue~\cite{Sarlin20cvpr} (SG), which uses a SfM point cloud for relocalization.
We also include results for SceneSqueezer~\cite{scenesqueezer2022cvpr} which requires less memory but at the cost of greater localization error.}

\subsection{Relocalization Evaluation}
\label{sec:evaluation}
\diff{Following the setup described in~\Cref{subsec:details}, }
we observe in~\Cref{tab:svfr_comparison_partial} an increase in the number of inliers, accompanied by a decrease in the rotation and translation errors as the camera pose approaches the (true) pose.
This observation, suggests that \nickname faithfully produces descriptors from poses seen during training as well as for new views. 

\diff{Given a query test image, we use DenseVLAD~\cite{torii2015densevlad} to retrieve the pose $\mathbf{\hat{T}}_q$ of the most similar image observed during training.}
This approach is also consistent with the methods used by~\cite{germain2022cvpr} and~\cite{moreau2023crossfire}.
\begin{table}[t]
    \begin{center}
        \setlength{\tabcolsep}{4pt}
        \begin{tabular}{l|c|c|c|c}
            \toprule
            Method & Prior Err. & Iter 1 & Iter 2 & Iter 3 \\
            \midrule
            Retrieval & 21.9, 12.13 & 0.8, 0.21 & 0.7, 0.19 & 0.7, 0.18 \\
            Constant & 147.6, 29.94 & 1.0, 0.28 & 0.7, 0.19 & 0.7, 0.18 \\
            \bottomrule
        \end{tabular}
    \end{center}%
    \vspace*{-2mm}
    \caption{\diff{\textbf{Different pose initialization priors.} We report the median localization errors $(cm, deg)$ at different iterations of the \pnpransac scheme on the chess scene of the 7-Scenes dataset. The pose initialization algorithms used are DenseVLAD for \textit{Retrieval} and the first pose of the training set for \textit{Constant}.}
    }
    \label{tab:pose_initialization_study}
    \vspace*{-4mm}
\end{table}

\diff{To evaluate the impact of poor initialization on localization, we used the first pose per scene from the 7-Scenes dataset~\cite{shotton13cvpr}  as a uniform prior for all images, instead of DenseVLAD~\cite{torii2015densevlad} pose priors.%
This simulates increasing noise as the query pose moves further from the initial pose. As shown in~\Cref{tab:pose_initialization_study} (more results in the supplementary), despite imprecise priors, the \pnpransac iterations reliably yield accurate localization, demonstrating the robustness of our method to initial pose variations.}

\begin{table*}[t]
	\centering
		\begin{tabular}{ll|ccccccc|c}
			\toprule
			 Category & Method & Chess & Fire &  Heads & Office & Pumpkin & Kitchen & Stairs & Average \\
			 & & \multicolumn{8}{c}{(cm, deg)} \\
			 \midrule
			\multirow{4}{*}{IBMs}  
			& PoseNet17~\cite{Kendall15iccv} & 13, 4.5 & 27, 11.3 & 17, 13.0 & 19, 5.6 & 26, 4.8 & 23, 5.4 & 35, 12.4 & 22.9, 8.1 \\
                & MapNet~\cite{brahmbhatt2018mapnet} & 8, 3.3 & 27, 11.7 & 18, 13.3 & 17, 5.2 & 22, 4.0 & 23, 4.9 & 30, 12.1 & 20.7, 7.8 \\
                & PAEs~\cite{shavit2022pae} & 12, 5.0 & 24, 9.3 & 14, 12.5 & 19, 5.8 & 18, 4.9 & 18, 6.2 & 25, 8.7 & 18.6, 7.5 \\
                & LENS~\cite{moreau2022PMLR} & 3, 1.3 & 10, 3.7 & 7, 5.8 & 7, 1.9 & 8, 2.2 & 9, 2.2 & 14, 3.6 & 8.3, 3.0 \\
                \midrule
			\multirow{1}{*}{HM}	
                & $\text{HLoc}^{\text{RGB-D}}_{\text{SP+SG}}$~\cite{sarlin2019coarse} & 2, 0.8 & 2. 0.8 & 1, 0.8	& 3, 0.8 & 4, 1.1 & 3, 1.1 & 4, 1.2 & 2.7, 0.9 \\
                \midrule
			\multirow{3}{*}{SBMs}
                & SC-WLS~\cite{wu2022scwls} & 3, 0.8 & 5, 1.1 & 3, 1.9 & 6, 0.9 & 8, 1.3 & 9, 1.4 & 12, 2.8 & 6.6, 1.5 \\
			& DSAC* (RGB)~\cite{brachmann2021dsacstar} & 2, 1.1 & 2, 1.2 & 1, 1.8 & 3, 1.2 & 4, 1.3 & 4, 1.7 & 3, 1.2 & 2.7, 1.4 \\
                & ACE~\cite{brachmann2023ace} & 2, 1.1 & 2, 1.8 & 2, 1.1 & 3, 1.4 & 3, 1.3 & 3, 1.3 & 3, 1.2 & 2.6, 1.3 \\
			\midrule
			\multirow{9}{*}{SFRMs}
			& FQN~\cite{germain2022cvpr} & 4, 1.3 & 5, 1.8 & 4, 2.4 & 10, 3.0 & 9, 2.5 & 16, 4.4 & 140, 34.7 & 27.4, 7.4 \\
                & CROSSFIRE~\cite{moreau2023crossfire} & 1, 0.4 & 5, 1.9 & 3, 2.3 & 5, 1.6 & 3, 0.8 & 2, 0.8 & 12, 1.9 & 4.4, 1.4 \\
                & NeRF-loc~\cite{liu2023nerfloc} & 2, 1.1 & 2, 1.1 & 1, 1.9 & 2, 1.1 & 3, 1.3 & 3, 1.5 & 3, 1.3 & 2.3, 1.3 \\
                \cmidrule{2-10}
                & (Ours) \nickname\textsubscript{Alike-t} & 1, 0.3 & 1, 0.5 & 1, 0.4 & 2, 0.6 & 2, 0.4 & 1, 0.3 & 4, 1.1 & \doubleunderline{1.7, 0.5} \\
                & (Ours) \nickname\textsubscript{Alike-s} & 1, 0.2 & 2, 0.6 & 1, 0.4 & 2, 0.4 & 1, 0.3 & 4, 0.9 & 5, 1.5 & 2.3, 0.6 \\
                & (Ours) \nickname\textsubscript{Alike-n} & 1, 0.2 & 1, 0.4 & 1, 0.6 & 2, 0.4 & 1, 0.3 & 1, 0.3 & 6, 1.6 & 1.9, 0.5 \\
                &  (Ours) \nickname\textsubscript{Alike-l} & 1, 0.2 & 1, 0.3 & 1, 0.4 & 2, 0.4 & 1, 0.3 & 1, 0.2 & 3, 0.8 & \textbf{1.4, 0.4}\\
			& (Ours) \nickname\textsubscript{SP} & 1, 0.2 & 1, 0.4 & 1, 0.3 & 2, 0.4 & 1, 0.3 & 1, 0.2 & 4, 1.0 & \underline{1.6, 0.4}\\
			\bottomrule
		\end{tabular}
    \caption{\textbf{6-DoF median localization errors on the 7-Scenes dataset~\cite{shotton13cvpr}}. Comparison of visual localization methods. 
        The overall top three results are shown in \textbf{bold}, \underline{underline}, and \doubleunderline{double-undeline}.}
        \label{tab:evaluation_pose_7scenes}
\end{table*}

\noindent\textbf{Indoor: 7-Scenes.} The 7-Scenes\cite{shotton13cvpr} dataset includes a sequence of frames recorded with a Kinect RGB-D camera, of seven different indoor environments. 
The data presents motion blur, small baselines between the frames, and occlusion, all typical elements for real-life indoor robotics applications. 
We use the full image size of 640$\times$480 pixels for training and testing.
For evaluation, we used the poses provided by~\cite{Brachmann2021ICCV} as ground truth.
In~\Cref{tab:evaluation_pose_7scenes}, we report the median translation and rotation errors (in centimetres and degrees) for state-of-the-art (SOTA) pose relocalization methods, divided into categories according to the data representation paradigm.
Our system outperforms the SOTA SFR methods by 39\% and 69\% for the translational and rotational errors, respectively.
\nickname also yields better results than the SBMs, IBMs, and HLoc~\cite{sarlin2019coarse} approaches.
In particular, for the indoor dataset, we report the results from the implementation repository of HLoc~\cite{sarlin2019coarse}, which are obtained using the depth maps provided by~\cite{brachmann2021dsacstar}.
We limited the number of landmarks for the indoor environments to 1,500.

\begin{table*}[ht]
	\centering
		\begin{tabular}{ll|ccccc|cc}
			\toprule
			Category & Method & College & Court & Hospital & Shop & Church & Average & Average\\
			& & \multicolumn{6}{c}{(cm, deg)} & w/o Court \\
			\midrule
			\multirow{4}{*}{IBMs} 
			& PoseNet~\cite{Kendall15iccv} & 88, 1.0 & 683, 3.5 & 88, 3.8 & 157, 3.3 & 320, 3.3 & 267, 3.0 & 163, 2.9\\
                & MapNet~\cite{brahmbhatt2018mapnet} & 107, 1.9 & 785, 3.8 & 149, 4.2 & 200, 4.5 & 194, 3.9 & 287, 3.7 & 163, 3.6\\
                & PAEs~\cite{shavit2022pae} & 90, 1.5 & - & 207, 2.6 & 99, 3.9 & 164, 4.2 & - & 140, 3.1\\
                & LENS~\cite{moreau2022PMLR} & 33, 0.5 & - & 44, 0.9 & 27, 1.6 & 53, 1.6 & - & 39, 1.2\\
                \midrule
                \multirow{2}{*}{HM}
                & HLoc\textsubscript{SP+SG}~\cite{sarlin2019coarse} & 6, 0.1 & 10, 0.1 & 13, 0.2 & 3, 0.1 & 4, 0.1 & \textbf{7, 0.1} & 
                \textbf{7, 0.1}\\
                & \diff{SceneSqueezer~\cite{scenesqueezer2022cvpr}} & \diff{27, 0.4} & - & \diff{37, 0.5} & \diff{11, 0.4} & \diff{15, 0.4} & - & \diff{23, 0.4}\\
                \midrule
                \multirow{3}{*}{SBMs}
                & SC-WLS~\cite{wu2022scwls} & 14, 0.6 & 164, 0.9 & 42, 1.7 & 11, 0.7 & 39, 1.3 & 54, 0.7 & 27, 1.1 \\
                & DSAC* (RGB)~\cite{brachmann2021dsacstar} & 18, 0.3 & 34, 0.2 & 21, 0.4 & 5, 0.3 & 15, 0.6 & 19, 0.3 & 15, 0.4 \\
                & ACE~\cite{brachmann2023ace} & 28, 0.4 & 42, 0.2 & 31, 0.6 & 5, 0.3 & 19, 0.6 & 25, 0.4 & 21, 0.5\\
                \midrule
                \multirow{9}{*}{SFRMs}
                & FQN-MN~\cite{germain2022cvpr} & 28, 0.4 & 4253, 39.2 & 54, 0.8 & 13, 0.6 & 58, 2.0 & 881, 8.6 & 38, 1.0\\
                & CROSSFIRE~\cite{moreau2023crossfire} & 47, 0.7 & - & 43, 0.7 & 20, 1.2 & 39, 1.4 & -  & 37, 1.0\\
                & NeRF-loc~\cite{liu2023nerfloc} & 11, 0.2 & 25, 0.1 & 18, 0.4 & 4, 0.2 & 7, 0.2 & \underline{13, 0.2} & \underline{10, 0.3} \\
                \cmidrule{2-9}
			& (Ours) \nickname\textsubscript{Alike-t} & 17, 0.3 & 29, 0.1 & 20, 0.4 & 5, 0.3 & 11, 0.4 & 16, 0.3 & 13, 0.4 \\
			& (Ours) \nickname\textsubscript{Alike-s} & 16, 0.2 & 32, 0.2 & 21, 0.4 & 6, 0.3 & 11, 0.4 & 17, 0.3 & 14, 0.4 \\
			& (Ours) \nickname\textsubscript{Alike-n} & 18, 0.3 & 32, 0.2 & 21, 0.4 & 5, 0.2 & 11, 0.3 & 17, 0.3 & 14, 0.3 \\
			& (Ours) \nickname\textsubscript{Alike-l} & 15, 0.2 & 27, 0.1 & 19, 0.4 & 5, 0.3 & 10, 0.3 & \doubleunderline{15, 0.3} & \doubleunderline{12, 0.3} \\
			& (Ours) \nickname\textsubscript{SP} & 18, 0.3 & 29, 0.2 & 27, 0.5 & 5, 0.3 & 11, 0.4 & 18, 0.3 & 15, 0.4 \\
			\bottomrule
		\end{tabular}
\caption{\textbf{6-DoF median localization errors on the Cambridge dataset~\cite{Kendall15iccv}}. Comparison of visual localization methods. 
        The overall top three results are shown in \textbf{bold}, \underline{underline}, and \doubleunderline{double-undeline}. Some NeRF-based methods fail when trained on the Great Court scene due to the poor image quality, whereas we can still track and train our voxel-based representation.}
    \label{tab:ablation_pose_cambridge}
    \vspace{-4mm}
\end{table*}

\noindent\textbf{Outdoor: Cambridge Landmarks.} The Cambridge Landmarks~\cite{Kendall15iccv} dataset includes images from large outdoor scenes that are 875 to 5,600 $\text{m}^2$ in size, with significant view and appearance changes. 
We reduce the size of the images to 512$\times$288 pixels for training and testing\diff{, since Alike~\cite{Zhao2022ALIKE} is faster when using lower-resolutions}.
\diff{For testing on the outdoor data set, we optimized 5,000 voxels for the Shop and Hospital scenes, and 10,000 for the others, based on scene size.}
As shown in~\Cref{tab:ablation_pose_cambridge}, similar to the indoor environment results, our system outperforms the SB and IB methods, with the hybrid methods achieving the best performance in large scenarios.
\diff{For the evaluation of HLoc~\cite{sarlin2019coarse}, we report the results from LIMAP~\cite{liu2023limap}, which, unlike the original HLoc, account for the camera lens distortion model when reprojecting the landmarks, providing the best results for outdoor environments.}
\nickname produces the second-best camera pose estimation results in the SFR category.
In particular, \nickname yields better results than FQN~\cite{germain2022cvpr} and CROSSFIRE~\cite{moreau2023crossfire} for all of the scenarios, while NeRF-loc~\cite{liu2023nerfloc} consistently provides an error of 1-3 centimeters lower than \nickname.
One of the main reasons for the reduced performance relative to NeRF-loc~\cite{liu2023nerfloc} is the inaccurate estimate of the position of the landmarks computed during the triangulation step. %
\diff{Unlike NeRF-loc~\cite{liu2023nerfloc}, which utilizes a fully trained NeRF model and RGB-D images for accurate 3D scene rendering, we adopt a sparse visual localization front-end that triangulates landmarks using tracks, which could %
introduce inaccuracies.}
\diff{The pose estimation errors observed with \nickname in the outdoor dataset may result from insufficient viewpoint variation, which makes accurate triangulation challenging given the large scene depth.}
However, despite the lower accuracy of \nickname for outdoor scenes, our method \diff{with Alike}, as analyzed in~\Cref{sec:memory_and_time_comparison}, uses less memory than~\cite{liu2023nerfloc} and does not require a complete neural rendering of the scene.
\subsection{Memory Usage and Training Complexity}
\label{sec:memory_and_time_comparison}
\begin{table}[t]
    \begin{center}
        \begin{tabular}{l|c|c}
            \toprule
            \multirow{1}{*}{Model} & 7-Scenes~\cite{shotton13cvpr} & Cambridge~\cite{Kendall15iccv} \\
            \midrule
            Posenet~\cite{Kendall15iccv} & 50MB & 50MB \\
            \midrule
            DSAC*~\cite{brachmann2021dsacstar} & 28MB & 28MB \\
            ACE~\cite{brachmann2023ace} & 4MB & 4MB \\
            \midrule
            \makecell[l]{HLoc\textsubscript{SP+SG}~\cite{sarlin2019coarse}\\(features only)} & \textgreater~4.2 GB & \textgreater~2.5 GB \\
            \midrule
            FQN~\cite{germain2022cvpr} & 2MB & 2MB \\
            CROSSFIRE~\cite{moreau2023crossfire} & 50MB & 50MB \\
            \makecell[l]{NeRF-loc~\cite{liu2023nerfloc}\\(backbone only)} & \textgreater~129 MB & \textgreater~129 MB\\
            \midrule
            Ours (Alike-t) & 13MB & 83MB \\
            Ours (Alike-l) & 19MB & 128MB \\
            Ours (SuperPoint) & 32MB & 216MB \\
            \bottomrule
        \end{tabular}
    \end{center}%
    \vspace*{-2mm}
    \caption{\textbf{Scene representation size comparison}.
    Comparison between the size of the largest checkpoints obtained with \nickname. The Chess~\cite{shotton13cvpr} and King's College~\cite{Kendall15iccv} scenes result in larger checkpoints than the other scenes. 
    For HLoc~\cite{sarlin2019coarse}, we report the memory used to store the features only in the two scenes.
    }
    \label{tab:memory_and_time_comparison}
\end{table}

In~\Cref{tab:memory_and_time_comparison}, we compare the checkpoint memory requirements of \nickname for both datasets against other methods.
\nickname uses more memory than FQN~\cite{germain2022cvpr} as our method represents each landmark independently.
However, this enables our representation to scale effectively, rendering high-dimensional descriptors in large scenes and consistently outperforming FQN~\cite{germain2022cvpr}. 
On the other hand, \nickname uses less than half the memory required by CROSSFIRE~\cite{moreau2023crossfire}. 
However, for large outdoor scenes, our method does require more memory. 
Despite the larger memory footprint, \nickname delivers consistently better localization performance than CROSSFIRE~\cite{moreau2023crossfire}.
\nickname\textsubscript{Alike-l} also requires less memory compared to NeRF-loc~\cite{liu2023nerfloc} \diff{for indoor and outdoor scenes.}

Our approach also performs favorably compared to hybrid methods.
For instance, \nickname uses over 200 times less memory than HLoc~\cite{sarlin2019coarse} with SP+SG~\cite{detone2018superpoint, Sarlin20cvpr} in indoor scenes. 
\diff{Methods like SceneSqueezer~\cite{scenesqueezer2022cvpr} significantly reduce the memory requirements (0.3 MB) for HBs, but at the cost of increased localization error.}
Image-based methods generally require less memory than structure-based ones but with the caveat that they provide less accurate pose estimates in general.
\diff{Advantages of our method include its ability to train a voxel from feature tracks, making it suitable for robotics localization pipelines.}
\subsection{Rendering Capabilities}
\label{sec:rendering_capabilities}
When extracting feature points from an image (i.e., keypoints and associated descriptors), the descriptors must be largely view-invariant to ensure robust matching from different viewpoints.
\diff{To evaluate the view-invariance of feature descriptors, we extract dense descriptor maps from a sequence of images taken from different angles. For each image, we use Alike-l~\cite{Zhao2022ALIKE} to extract features from a target image and compute similarity scores between each feature and the dense maps. The same process is applied to descriptors rendered by \nickname, but using the ground truth pose for rendering.}

\begin{figure}[t]
    \centering
    \includegraphics[width=\linewidth, trim=0in 0in 0.5in 0.2in, clip]{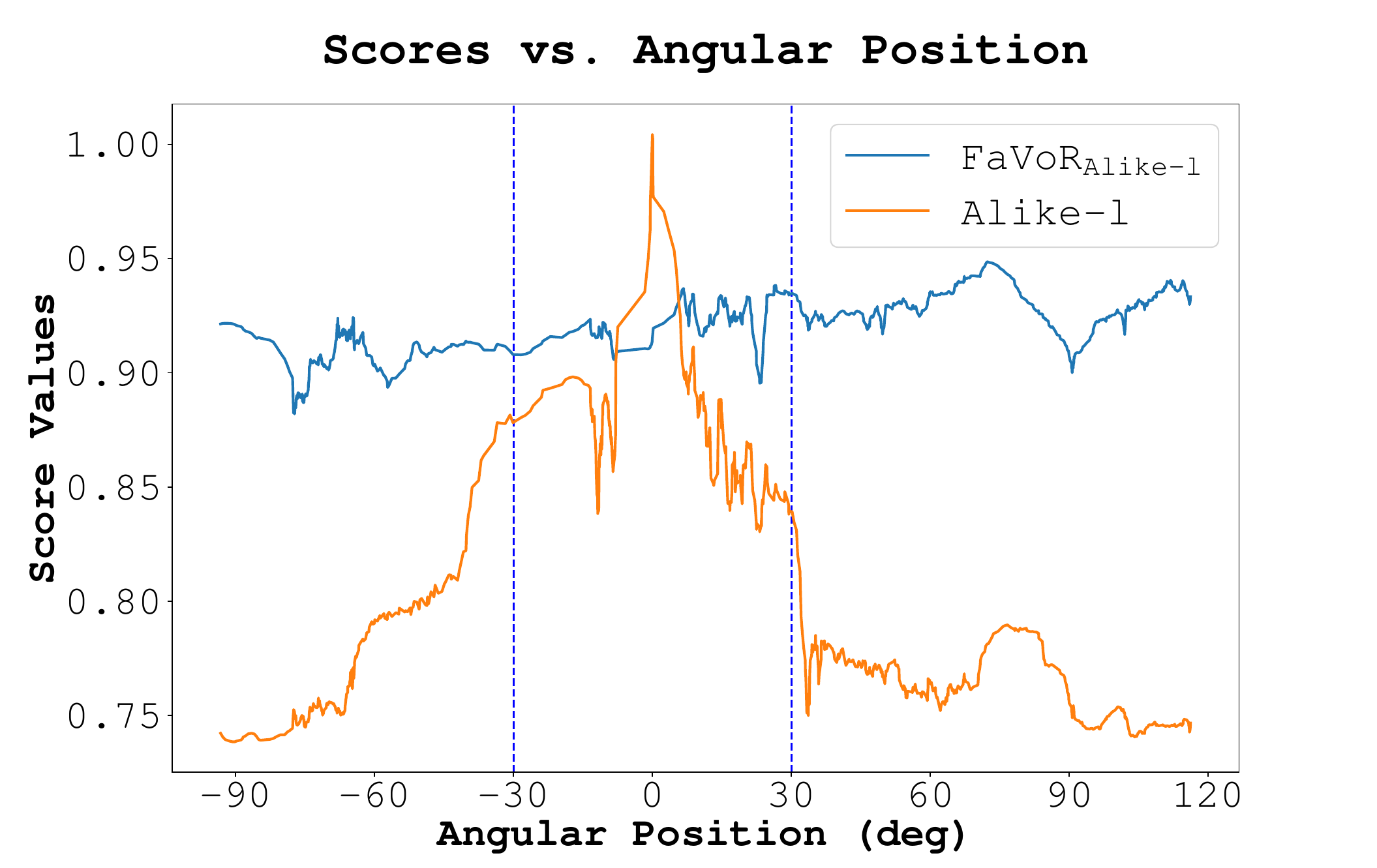}
    \caption{\textbf{Median similarity score versus viewing angle.} In blue is the smoothed median score for \nickname\textsubscript{Alike-l} obtained by convolving the descriptors rendered at different view angles with the corresponding dense descriptors map of each query image. In orange is the smoothed median score of Alike-l features extracted from the starting image (at angle 0 deg) convolved with the subsequent images in the test sequence.
    }
    \label{fig:median_score_graph_superpoint}
\end{figure}

\diff{\Cref{fig:median_score_graph_superpoint} shows the median values of the top thirty similarity scores per image at different query angles}, for Alike-l and \nickname trained to render Alike-l features.
\nickname maintains an almost constant value, indicating that the descriptors are rendered faithfully from unseen views close to the ones observed during training.
In contrast, there is a noticeable degradation in the median similarity scores when attempting to match features directly extracted by Alike-l, with a significant drop occurring beyond $\pm 30$ degrees.
\Cref{fig:median_score_graph_superpoint} demonstrates the effectiveness of \nickname in maintaining descriptor rendering fidelity across varying viewpoints.

\section{Conclusion}
\label{sec:conclusion}

In this paper, we presented \nickname, a novel, globally sparse and locally dense vision-based camera relocalization method.
Our approach involves tracking features over a sequence of RGB images with known poses and learning a 3D descriptor representation that models variations in appearance according to viewpoint. %
Our method is also suitable for integration with existing track-based localization systems for robotics.

There are several directions for future work. 
The voxels in \nickname are independent of each other, allowing for versatile training approaches. Although our current implementation optimizes each voxel sequentially, parallelizing this process could significantly speed up training, potentially approaching the speed of training a single voxel.
We plan to implement these parallelization techniques in the future to enhance efficiency.
To increase robustness against occlusions, additional modifications could involve utilizing full-image descriptors for landmark retrieval.


{\small
\bibliographystyle{ieee_fullname}
\bibliography{ref}
}


    \appendix
    \section*{Supplementary Material}
    
We report on the results obtained by \nickname at various iterations of the PnP-RANSAC scheme in in~\Cref{sec:more_results}.
In~\Cref{sec:psnr_model_size}, we discuss the tradeoff between the voxel resolution and both the rendering capabilities and matching performance of our system.
We also report the similarity responses for the Cambridge Landmarks~\cite{Kendall15iccv} dataset, discussing the evidence of a lack of accuracy in the landmark triangulation in~\Cref{sec:score_response_suppl}. 
Finally, we provide more details on our training losses and our landmark triangulation method in Appendices~\ref{sec:trainign_losses} and~\ref{sec:landmarks_traingulation}, respectively.

\section{Extended Analysis of \nickname Performance and Error Computation}
\label{sec:more_results}
In this section, we report the pose estimation errors obtained with different feature extractors at various iterations of the iterative PnP-RANSAC scheme.
Specifically, we report the values for Alike-t, Alike-s, Alike-n, Alike-l~\cite{Zhao2022ALIKE} and SuperPoint~\cite{detone2018superpoint} with 64, 94, 128, 128, and 256 channels descriptors, respectively.

\Cref{tab:svfr_comparison_7scenes} for the 7-Scenes~\cite{shotton13cvpr} and~\Cref{tab:svfr_comparison_cambridge} for the Cambrdige~\cite{Kendall15iccv} datasets give the median pose estimate at the 1st, 2nd, and 3rd iterations of PnP-RANSAC, and the respective average number of inlier points per image (used to compute the pose estimate). 
The tables also report the success rates of the PnP-RANSAC iterative scheme at the various iterations, i.e., the ratio between the number of successful estimates and the total number of queries.
The data shows a clear trend. Namely, the average number of inliers per image increases as the estimated camera pose converges towards the true query image pose (i.e., as the pose estimate error decreases).
Furthermore, although there is a difference in matching performance between the various Alike networks (as reported in the Alike~\cite{Zhao2022ALIKE} manuscript), our descriptor representation effectively `flattens' these differences, enabling robust matching despite view changes.

\section{PSNR versus Voxel Resolution}
\label{sec:psnr_model_size}
\begin{figure}[ht]
    \centering
    \includegraphics[width=\linewidth]{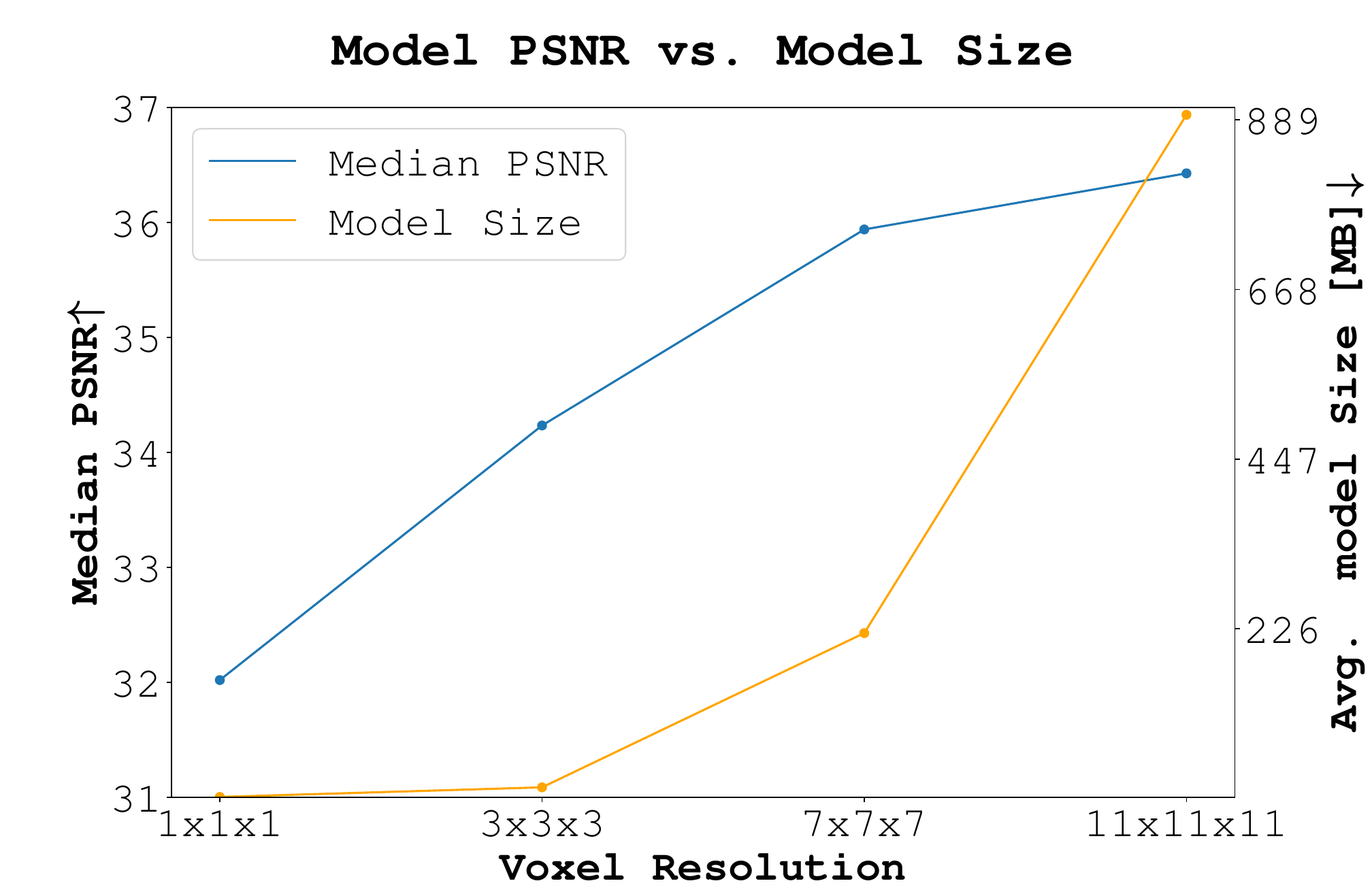}
    \vspace{-2ex}
    \caption{\textbf{PSNR and model size versus grid resolution.} We report the median peak signal-to-noise ratio (PSNR) and the average checkpoint size for \nickname\textsubscript{Alike-l} at different grid resolutions of the voxel representation. }
    \label{fig:psnr_vs_modelsize}
\end{figure}

\begin{figure}[ht]
    \centering
    \includegraphics[width=\linewidth]{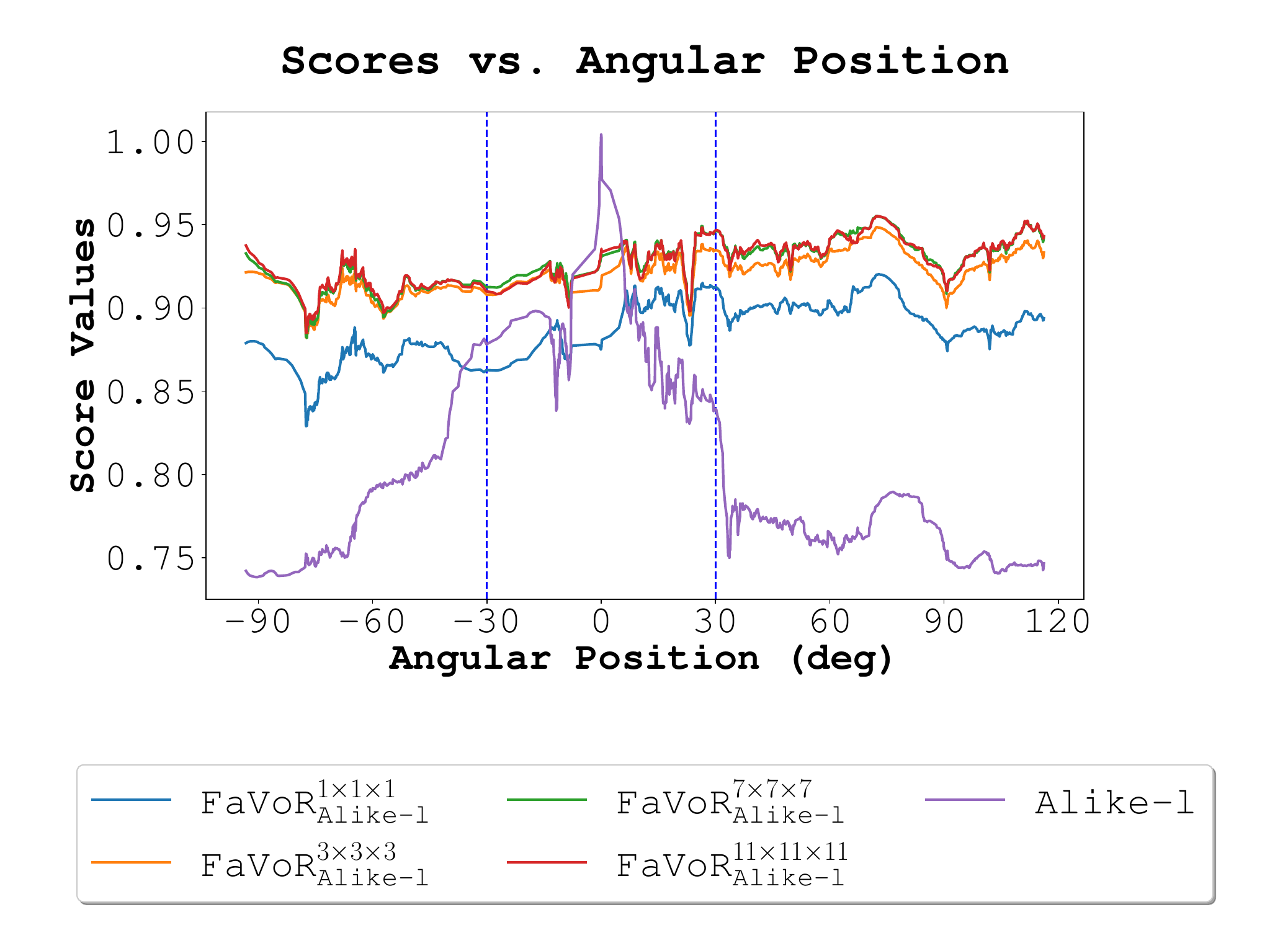}
    \vspace{-2ex}
    \caption{\textbf{Similarity response scores versus grid resolution at different view angles.} We compare the different grid resolutions' capacity to provide high score similarity score results at different view angles. Higher scores lead to better matching meaning that the rendered descriptors properly match the appearance of the ones extracted by Alike-l~\cite{Zhao2022ALIKE}. }
    \label{fig:scores_vs_grid}
\end{figure}

The number of sub-voxels in each voxel representing a landmark determines the grid resolution, $R \times R \times R$.
The grid resolution directly impacts the rendering quality of the descriptor patches, increasing or decreasing the peak signal-to-noise ratio (PSNR) values.
The PSNR calculation is given by
\begin{align}
    \text{PSNR} = 10\log_{10}\left(\frac{\text{MAX}^2}{\text{MSE}}\right),
\end{align}
where MAX = 2 is the maximum span of the descriptor values, i.e, in the range (-1, 1), and MSE is the mean squared error between the ground-truth patch and the rendered patch. 
A grid resolution of $R \times R \times R$ implies that the grid contains $R \cdot R \cdot R$ nodes, where each node (vertex) encodes $C$ channels (equal to the number of channels of the descriptor). The chosen grid resolution impacts the overall model size.

The plot in~\Cref{fig:psnr_vs_modelsize} shows the median PSNR values at different grid resolutions and the corresponding model size on the chess scene of the 7-Scenes dataset~\cite{shotton13cvpr}.
The graph shows that beyond a certain grid resolution, the improvement in terms of PSNR is decreasing while, in contrast, the model size grows exponentially.
Therefore, as a tradeoff between model size and \textit{good} rendering capabilities of our representation, we choose a grid resolution of 3 $\times$ 3 $\times$ 3 for our model.
Also, this resolution choice is supported by the median score values, obtained as described in the manuscript in Section 4.4, reported in~\Cref{fig:scores_vs_grid}.
~\Cref{fig:scores_vs_grid} shows the median score values obtained in the chess scene of the 7-Scenes dataset~\cite{shotton13cvpr} with Alike-l~\cite{Zhao2022ALIKE}, at different grid resolutions.
We note that grid resolutions greater than 1 $\times$ 1 $\times$ 1 yield a similar score response; hence, we choose the lower resolution to save on memory.

    \section{Score Response}
\label{sec:score_response_suppl}

\begin{figure*}[ht]
    \centering
    \begin{subfigure}[b]{0.33\linewidth}
        \includegraphics[width=\textwidth]{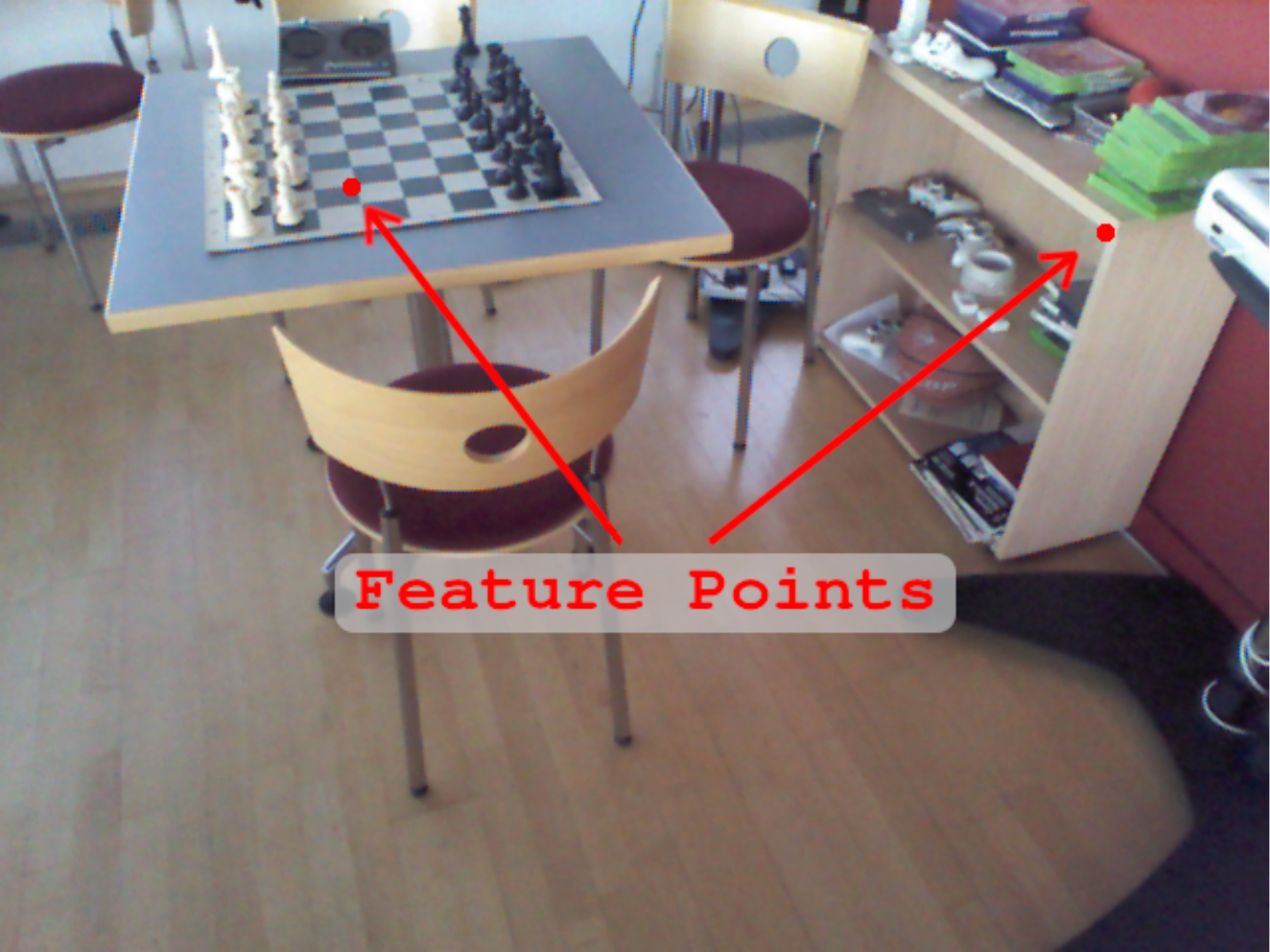} 
        \caption{}
        \label{fig:short-ac}
    \end{subfigure}
    \begin{subfigure}[b]{0.33\linewidth}
        \includegraphics[width=\textwidth]{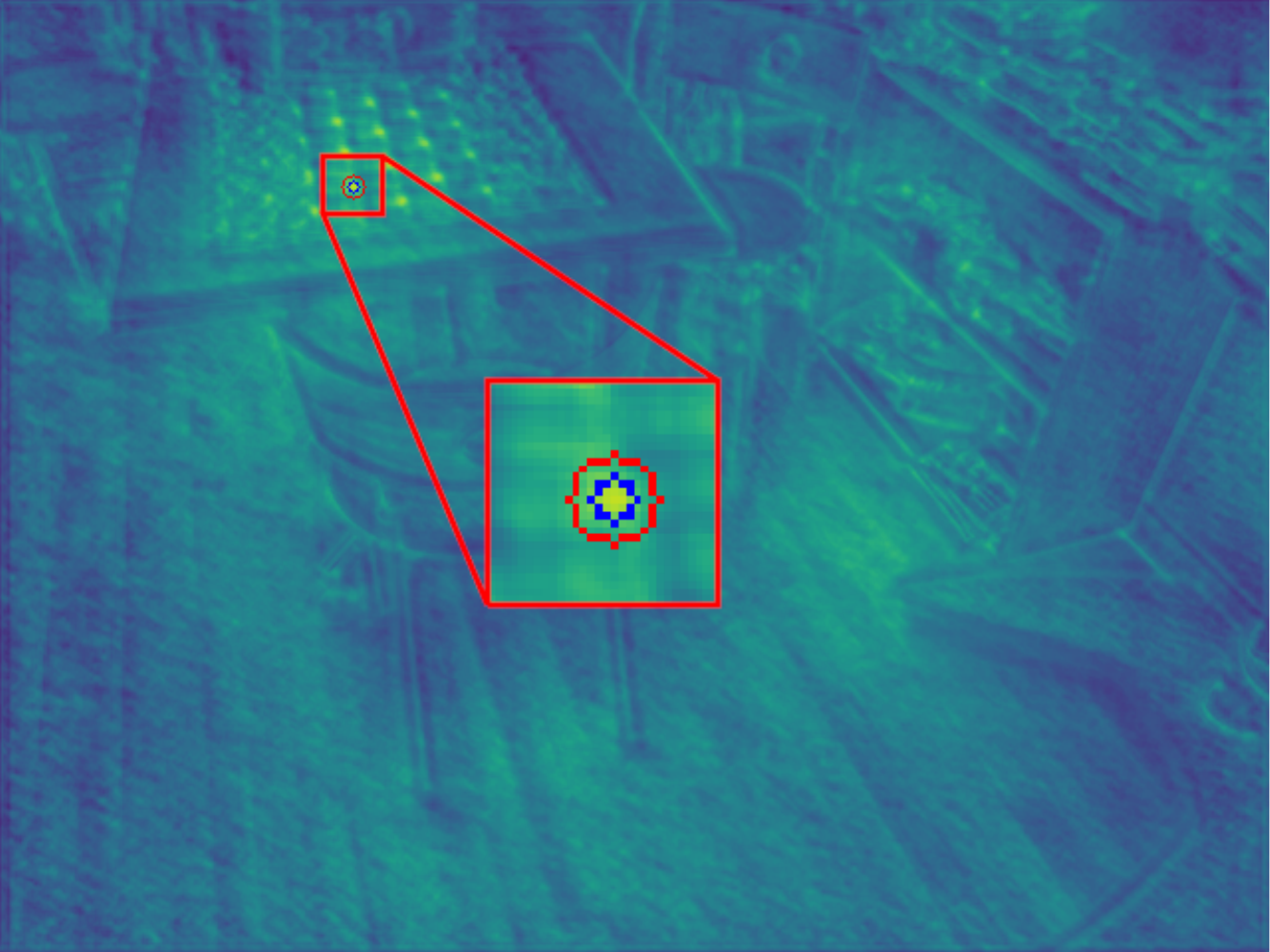} 
        \caption{}
        \label{fig:short-bc}
    \end{subfigure}
    \begin{subfigure}[b]{0.33\linewidth}
        \includegraphics[width=\textwidth]{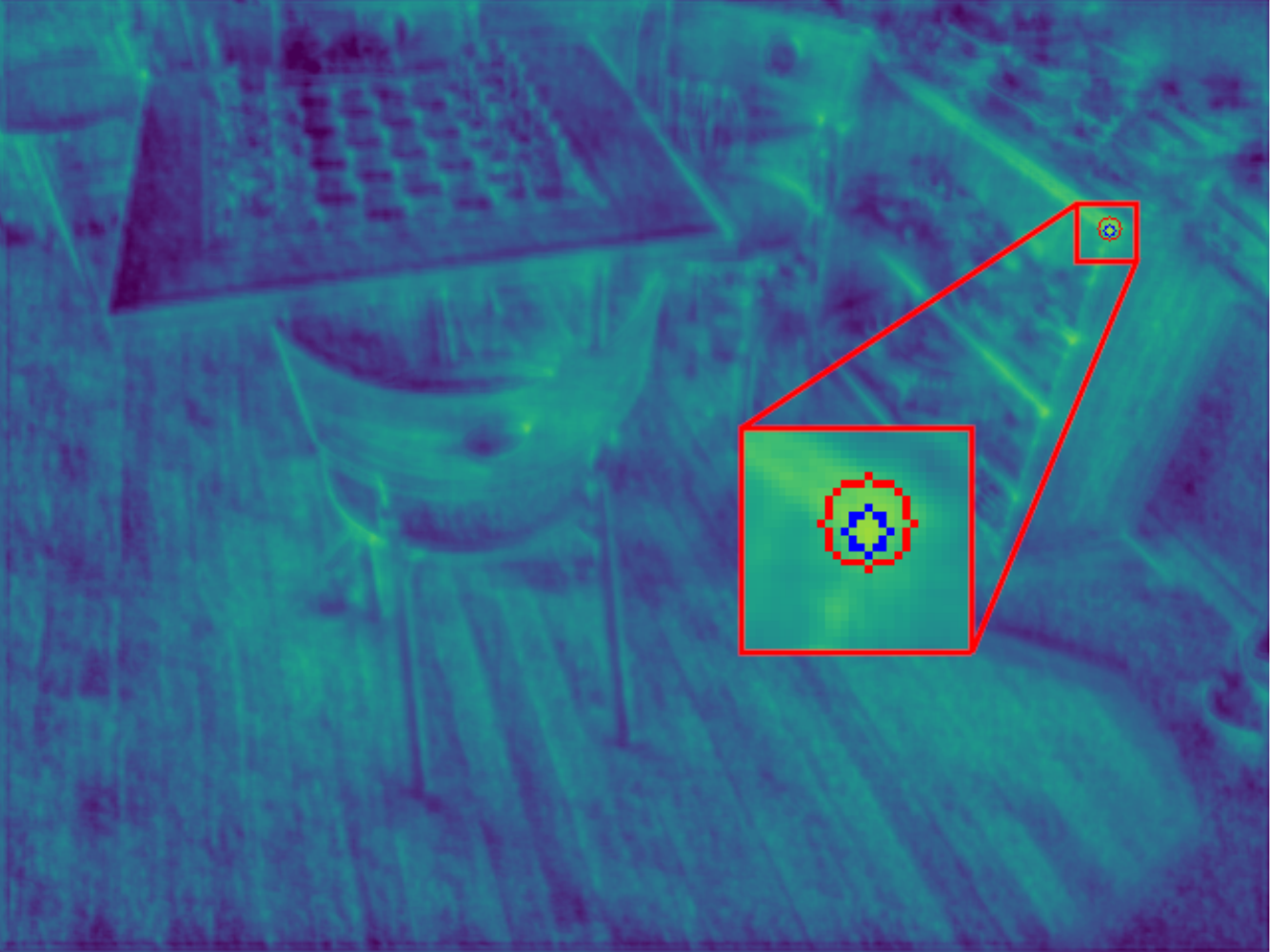} 
        \caption{}
        \label{fig:short-cc}
    \end{subfigure}
    \begin{subfigure}[b]{0.33\linewidth}
        \includegraphics[width=\textwidth]{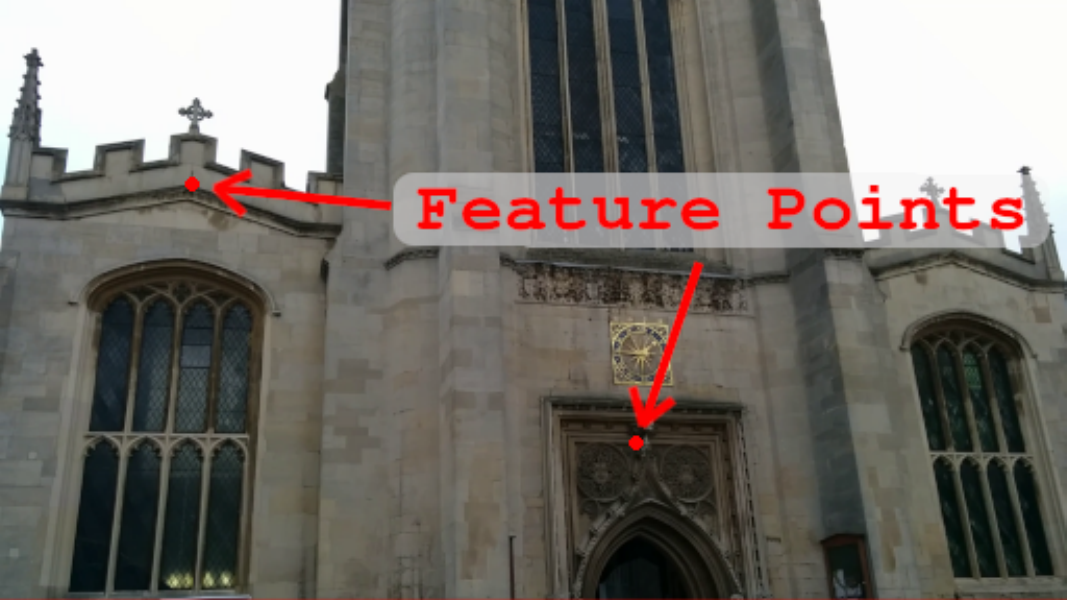} 
        \caption{}
        \label{fig:short-a1}
    \end{subfigure}
    \begin{subfigure}[b]{0.33\linewidth}
        \includegraphics[width=\textwidth]{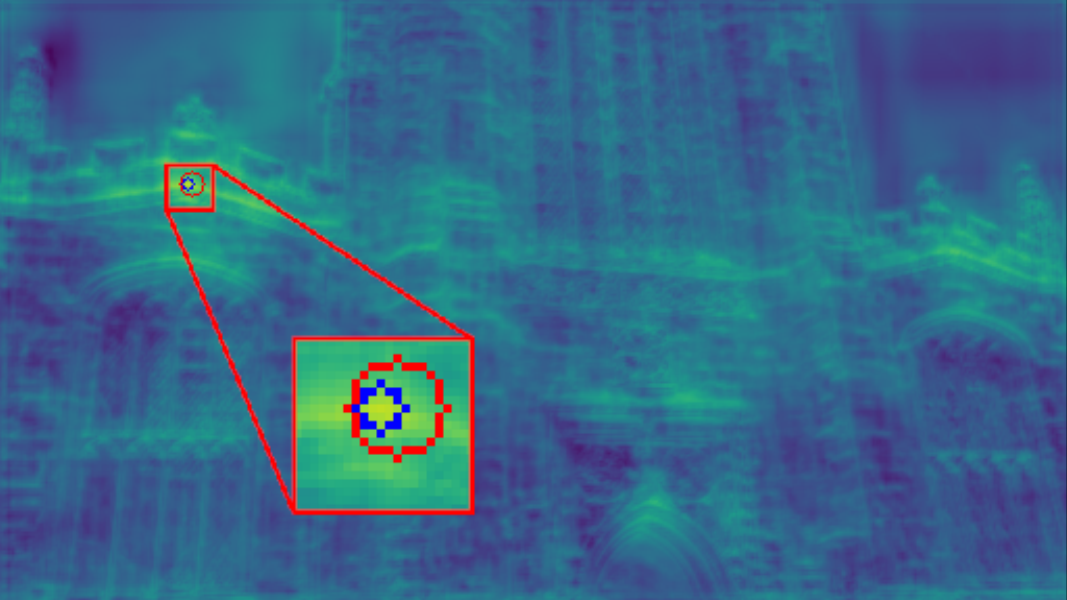} 
        \caption{}
        \label{fig:short-b1}
    \end{subfigure}
    \begin{subfigure}[b]{0.33\linewidth}
        \includegraphics[width=\textwidth]{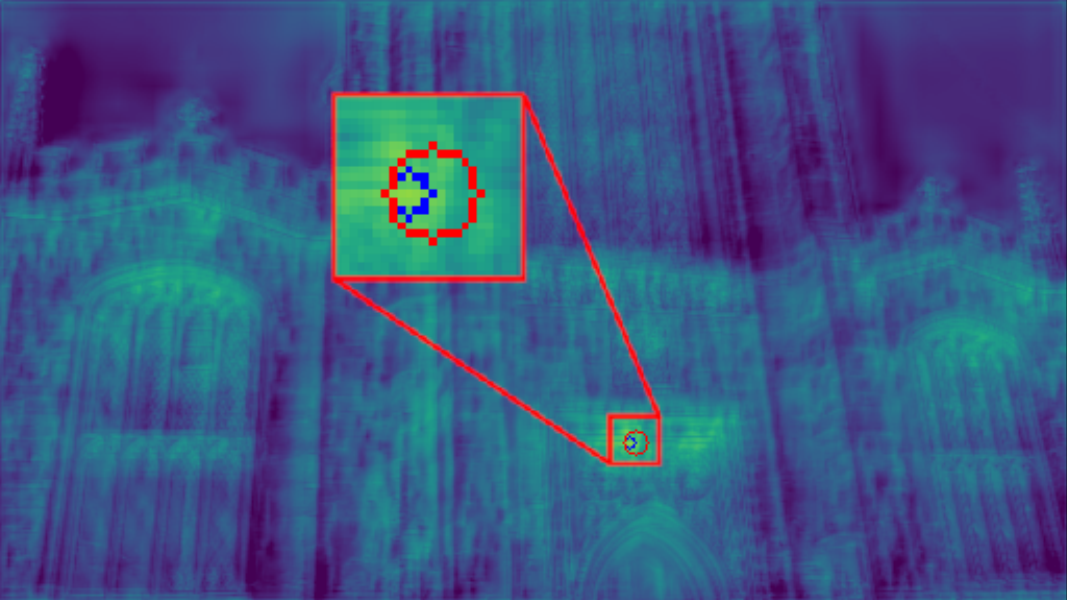} 
        \caption{}
        \label{fig:short-c1}
    \end{subfigure}
    \caption{\label{fig:feature_map_matching_chess_stmary}\textbf{Visualization of similarity response.}  We render a feature tracked during training using the Alike-l descriptor from an unseen view on the 7-Scenes~\cite{shotton13cvpr} and the Cambridge Landmarks~\cite{Kendall15iccv} datasets. On the left, a) and d) display the ground truth positions of the rendered feature points, obtained by projecting the triangulated landmarks on the camera plane, in red. At the same time, b), e) and c), f) show the similarity response between the rendered features and the target image dense descriptor map. The yellow color indicates a strong response, concentrated around the feature positions shown in a), demonstrating the effectiveness of our descriptor rendering approach. The small circle in blue is the circle center at the highest score response, the red circle is centered at the project landmark position.
    }
\end{figure*}

In~\Cref{fig:feature_map_matching_chess_stmary} we show the response score map obtained by convolving the dense descriptor map extracted using Alike-l~\cite{Zhao2022ALIKE} with two descriptors rendered using \nickname\textsubscript{Alike-l}. 
In particular, we draw a red circle centred at the projection of the triangulated landmarks on the camera plane.
We add another circle (in blue) that is centred at the coordinates of the pixel with the strongest similarity response.
Both the circles should be concentric to provide an accurate pose estimate.
However, we notice that, most of the time, the circles do not have the same centre for the samples obtained from the Cambridge Landmarks dataset~\cite{Kendall15iccv}.
This misalignment may be due to an imprecise triangulation of the landmarks, given the depth uncertainty for the large scenes of the Cambridge Landmarks dataset~\cite{Kendall15iccv}.

    \section{Training and Losses}
\label{sec:trainign_losses}

\begin{figure}[t]
    \centering
    \begin{subfigure}[b]{0.3\linewidth}
        \includegraphics[width=\textwidth]{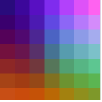} 
        \caption{}
        \label{fig:short-ap}
    \end{subfigure}
    \begin{subfigure}[b]{0.3\linewidth}
        \includegraphics[width=\textwidth]{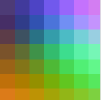} 
        \caption{}
        \label{fig:short-bp}
    \end{subfigure}
    \begin{subfigure}[b]{0.3\linewidth}
        \includegraphics[width=\textwidth]{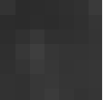} 
        \caption{}
        \label{fig:short-cp}
    \end{subfigure}    \caption{\label{fig:trained_patche_vs_gt}\textbf{Visualization of rendered vs ground truth descriptor patch.} We report the rendered descriptor patch a) and the corresponding ground-truth b) for Alike-l~\cite{Zhao2022ALIKE}, compressed from 128 channels to 3 using PCA for visualization purposes. The patch in c) represents the normalized difference between the rendered and the ground truth, darker is better.
    }
\end{figure}

To train our voxel representation of the descriptor patches, we used two main losses, the squared L2 norm loss and the cosine similarity loss.
We begin by training our model using the squared L2 norm and cosine similarity losses to enforce that the direction and norm of the rendered descriptors match the ground truth.
The cosine similarity loss is calculated as
\begin{align}
    \text{L}_{\cos}\left( \hat{\mathbf{d}}_{ij}^{uv}, \mathbf{d}_{ij}^{uv}\right) = 1 - \frac{\hat{\mathbf{d}}_{ij}^{uv}\cdot \mathbf{d}_{ij}^{uv}}{||\hat{\mathbf{d}}_{ij}^{uv}||\cdot||\mathbf{d}_{ij}^{uv}||},
\end{align}
where $\hat{\mathbf{d}}_{ij}^{uv}$ is the rendered descriptor and $\mathbf{d}_{ij}^{uv}$ is the one extracted by $\mathcal{F}$ from the patch $\mathbf{P}_{ij}$.
For the density grid, we use the cross-entropy loss, as in~\cite{Sun22cvpr}.
Also, to ensure a smooth representation of the descriptor patch, for the last 500 epochs, we introduce a total variation (TV) regularization term in the loss computation on the density and the descriptor parameters as described in~\cite{Sun22cvpr}.
The complete loss function for the voxel optimization is given by
\begin{multline}
\text{Loss}\left(\hat{\mathbf{d}}_{ij}^{uv}, \mathbf{d}_{ij}^{uv}\right) = \\ || \hat{\mathbf{d}}_{ij}^{uv} - \mathbf{d}_{ij}^{uv}||_2^2
    +\text{L}_{\cos}\left( \hat{\mathbf{d}}_{ij}^{uv}, \mathbf{d}_{ij}^{uv}\right)
    +\text{TV}.
\end{multline}
During training, we choose a learning rate that depends on the visibility of each sub-voxel.
In particular, we follow the same approach proposed by Sun~\etal~\cite{Sun22cvpr} where subvoxels visible from fewer views have a lower learning rate.
Choosing the learning rate according to the visibility of the voxels allows training to \textit{focus} more on the parts of the voxels that represent portions that have been observed from several views, and hence are more reliable than those with fewer observations.
\Cref{fig:trained_patche_vs_gt} shows a descriptor patch, the corresponding ground truth extracted using Alike-l~\cite{Zhao2022ALIKE}, both visualized using principal component analysis (PCA) to map the descriptor space to RGB colors, and the L2 norm between the two patches in the descriptor space.

    \section{Landrmark Triangulation}
\label{sec:landmarks_traingulation}

Our method requires landmarks positions to locate and train the associated voxels.
Vision-based localization systems, such as visual-inertial odometry or visual simultaneous localization and mapping, already provide a landmark's position in 3D space.
Hence, we opt for a simple-to-implement multi-view triangulation approach, since triangulation is not the main focus of our work.
Given a track containing $N$ poses $\mathbf{T}_i$, with $i=1 \hdots N$, and hence $N$ keypoints $\mathbf{k}_{ij}$ corresponding to the projection onto each camera plane of the landmark $\pmb{\ell}_j$, we wish to find the 3D coordinates $\prescript{\mathcal{W}}{}{\ell_j^x}, \prescript{\mathcal{W}}{}{\ell_j^y}, \prescript{\mathcal{W}}{}{\ell_j^z}$ in the world frame $\mathcal{W}$ of $\pmb{\ell}_j$.
An initial estimate of the coordinates  $\pmb{\ell}_j$ can be determined by two-view triangulation methods, such as the direct linear transform.
Following~\cite{Delaune20xvio}, we choose an anchor pose $\mathbf{T}_a$ from among the poses in the track, and express $\pmb{\ell}_j$ in camera coordinates for $\mathbf{T}_a$.
We define $\mathbf{T}_a \in \mathrm{SE}(3)$, and in general any $\mathbf{T}_i$, in terms of a rotation matrix $\mathbf{R}_a \in \mathrm{SO}(3)$ and a translation vector $\mathbf{t}_a \in \mathbb{R}^3$:
\begin{align}
    \begin{bmatrix}
        \prescript{\mathcal{A}}{}{\ell_j^x}\\
        \prescript{\mathcal{A}}{}{\ell_j^y}\\
        \prescript{\mathcal{A}}{}{\ell_j^z}
    \end{bmatrix} = \mathbf{R}_a^{T} 
    \begin{bmatrix}
        \prescript{\mathcal{W}}{}{\ell_j^x}\\
        \prescript{\mathcal{W}}{}{\ell_j^y}\\
        \prescript{\mathcal{W}}{}{\ell_j^z}
    \end{bmatrix}-\mathbf{R}_a^{T}\mathbf{t}_a .
    \label{eq:projection_anchor}
\end{align}
From~\Cref{eq:projection_anchor}, we can write the landmark coordinates in the world frame as a function of the anchor  pose coordinates:
\begin{align}
    \begin{bmatrix}
        \prescript{\mathcal{W}}{}{\ell_j^x}\\
        \prescript{\mathcal{W}}{}{\ell_j^y}\\
        \prescript{\mathcal{W}}{}{\ell_j^z}
    \end{bmatrix} = \mathbf{R}_a 
    \begin{bmatrix}
        \prescript{\mathcal{A}}{}{\ell_j^x}\\
        \prescript{\mathcal{A}}{}{\ell_j^y}\\
        \prescript{\mathcal{A}}{}{\ell_j^z}
    \end{bmatrix}+\mathbf{t}_a .
\end{align}
Hence, each time we need to determine the landmark coordinates $\pmb{\ell}_j$ in camera frame $\mathcal{T}_i$, associated with the pose $\mathbf{T}_i$ in the track, we can write:
\begin{align}
\allowdisplaybreaks
    \begin{bmatrix}
        \prescript{\mathcal{T}_i}{}{\ell_j^x}\\
        \prescript{\mathcal{T}_i}{}{\ell_j^y}\\
        \prescript{\mathcal{T}_i}{}{\ell_j^z}
    \end{bmatrix} & = \mathbf{R}_i^{T} \left(\mathbf{R}_a \begin{bmatrix}
        \prescript{\mathcal{A}}{}{\ell_j^x}\\
        \prescript{\mathcal{A}}{}{\ell_j^y}\\
        \prescript{\mathcal{A}}{}{\ell_j^z}
    \end{bmatrix}+\mathbf{t}_a\right) -\mathbf{R}_i^{T}\mathbf{t}_i \\[2mm]
    &= \mathbf{R}_i^{T} \mathbf{R}_a \begin{bmatrix}
        \prescript{\mathcal{A}}{}{\ell_j^x}\\
        \prescript{\mathcal{A}}{}{\ell_j^y}\\
        \prescript{\mathcal{A}}{}{\ell_j^z}
    \end{bmatrix}+ \mathbf{R}_i^{T}(\mathbf{t}_a-\mathbf{t}_i)
    \label{eq:land_in_i}
\end{align}
To improve the numerical stability of the optimization process, we represent $\begin{bmatrix}
    \prescript{\mathcal{A}}{}{\ell_j^x},
    \prescript{\mathcal{A}}{}{\ell_j^y},
    \prescript{\mathcal{A}}{}{\ell_j^z}
\end{bmatrix}^T$ using the inverse depth parametrization,
\begin{align}
    \alpha_j = \frac{\prescript{\mathcal{A}}{}{\ell_j^x}}{\prescript{\mathcal{A}}{}{\ell_j^z}},\quad
    \beta_j = \frac{\prescript{\mathcal{A}}{}{\ell_j^y}}{\prescript{\mathcal{A}}{}{\ell_j^z}},\quad
    \rho_j = \frac{1}{\prescript{\mathcal{A}}{}{\ell_j^z}}
\end{align}
We can then rewrite~\Cref{eq:land_in_i} as
\begin{align}
    \begin{bmatrix}
        \prescript{\mathcal{T}_i}{}{\ell_j^x}\\
        \prescript{\mathcal{T}_i}{}{\ell_j^y}\\
        \prescript{\mathcal{T}_i}{}{\ell_j^z}
    \end{bmatrix} = \frac{1}{\rho_j}\left(\mathbf{R}_i^{T} \mathbf{R}_a \begin{bmatrix}
        \alpha_j\\
        \beta_j\\
        1
    \end{bmatrix} + \rho_j\,\mathbf{R}_i^{T}(\mathbf{t}_a-\mathbf{t}_i)\right)
\end{align}
In turn, the camera measurement model is
\begin{align}
    \mathbf{\hat{z}}_{ij} &= \frac{1}{\prescript{\mathcal{T}_i}{}{\ell_j^z}}\begin{bmatrix}
        \prescript{\mathcal{T}_i}{}{\ell_j^x},
        \prescript{\mathcal{T}_i}{}{\ell_j^y}\\
        \end{bmatrix}^T,
\end{align}
where $\hat{\mathbf{z}}_{ij}$ are the normalized image plane coordinates of $\prescript{\mathcal{T}_i}{}{\ell_{j}}$.
The predicted measurement can be determined by transforming  $\mathbf{k}_{ij}$ into camera coordinates to obtain $\mathbf{z}_{ij}$. This involves back-projecting the keypoint coordinates $\mathbf{k}_{ij}$ from the image plane to the camera frame, followed by normalization,
\begin{align}
    \begin{bmatrix}
        x_{\mathbf{k}_{ij}}\\
        y_{\mathbf{k}_{ij}}\\
        z_{\mathbf{k}_{ij}}
    \end{bmatrix}&=\mathbf{K}^{-1}
    \begin{bmatrix}
        \mathbf{k}_{ij}\\
        1
    \end{bmatrix}\\
    \mathbf{z}_{ij} &= \frac{1}{z_{\mathbf{k}_{ij}}}\begin{bmatrix}
        x_{\mathbf{k}_{ij}},
        y_{\mathbf{k}_{ij}}\\
        \end{bmatrix}^T,
\end{align}
where $\mathbf{K}$ is the intrinsic camera calibration matrix.

Finally, we find $\alpha_j$, $\beta_j$, and $\rho_j$ via Levenberg-Marquardt optimization,
\begin{align}
    \mathbf{e}_{ij} &= \mathbf{z}_{ij} - \mathbf{\hat{z}}_{ij},\\
    u_{ij} &= \sqrt{\mathbf{e}_{ij}^T\mathbf{e}_{ij}},\\
    \rho(u) &= \frac{1}{2}\frac{c^2u^2}{c^2+u^2},\\
    \alpha_j^*,\beta_j^*,\rho_j^* = \argmin_{\alpha_j,\beta_j,\rho_j} &\sum_{i \in \mathcal{S}_j} \rho(u_{ij}(\mathbf{e}_{ij}(\mathbf{\hat{z}}_{ij}(\pmb{\ell}_{ij}(\alpha_j,\beta_j,\rho_j)))),
\end{align}
where $\rho(u)$ is a robust cost function~\cite{geman1986bayesian} parameterized by $c$, used to prevent outliers from negatively impacting the estimate of the landmark coordinates.

\section{Robustness to Pose Initialization Error}
\begin{table*}[!t]
	\centering
		\begin{tabular}{l|l|c|ccc}
			\toprule
			Scene & Method & Prior Err. & Iter 1 & Iter 2 & Iter 3 \\
			\midrule
                \multirow{2}{*}{chess} & Retrieval & 21.9, 12.13 & 0.8, 0.21 & 0.7, 0.19 & 0.7, 0.18 \\
                & Constant & 147.6, 29.94 & 1.0, 0.28 & 0.7, 0.19 & 0.7, 0.18 \\
                \midrule
                \multirow{2}{*}{fire} & Retrieval & 34.4, 13.2 & 0.8, 0.3 & 0.9, 0.4 & 0.8, 0.3 \\
                & Constant & 96.6, 35.6 & 1.2, 0.5 & 0.9, 0.4 & 0.8, 0.3\\
                \midrule
                \multirow{2}{*}{heads} & Retrieval & 15.8, 15.0 & 0.7, 0.5 & 0.7, 0.4 & 0.7, 0.4 \\
                & Constant & 45.7, 37.8 & 28.7, 17.2 & 2.5, 1.5 & 0.5, 0.4 \\
                \midrule
                \multirow{2}{*}{office} & Retrieval & 28.6, 11.1 & 1.7, 0.4 & 1.7, 0.4 & 1.6, 0.4 \\
                & Constant & 113.8, 67.4 & 158.3, 57.4 & 10.2, 2.5 & 1.7, 0.4 \\
                \midrule
                \multirow{2}{*}{pumpkin} & Retrieval & 31.4, 10.8 & 1.4, 0.3 & 1.4, 0.3 & 1.3, 0.3 \\
                & Constant & 137.0, 49.7 & 4.1, 1.1 & 1.5, 0.3 & 1.2, 0.2 \\
                \midrule
                \multirow{2}{*}{redkitchen} & Retrieval & 29.4, 12.0 & 1.2, 0.3 & 1.2, 0.3 & 1.2, 0.2 \\
                & Constant & 192.4, 39.3 & 7.8, 1.8 & 1.8, 0.4 & 1.1, 0.2 \\
                \midrule
                \multirow{2}{*}{stairs} & Retrieval & 26.2, 15.8 & 3.8, 1.0 & 3.0, 0.8 & 2.7, 0.8 \\
                & Constant & 178.9, 16.3 & 231.4, 34.2 & 106.8, 16.2 & 4.7, 1.32 \\
			\bottomrule
		\end{tabular}
    \caption{\diff{\textbf{Different pose initialization priors for 7-Scenes dataset~\cite{shotton13cvpr}}. We report the 6-DoF median pose errors $(cm, deg)$ obtained with \nickname\textsubscript{Alike-l} for different pose initialization methods. The results show that \nickname is robust to different initial poses as it converges to small localization errors after iterating the \pnpransac scheme. }}
        \label{tab:pose_initialization_study_full}
\end{table*}

In~\Cref{tab:pose_initialization_study_full} we report the 6-DoF median localization errors for the 7-Scenes~\cite{shotton13cvpr} dataset using two pose initialization methods: the first frame of the test sequence (Constant) and DenseVLAD~\cite{torii2015densevlad} (Retrieval).
We perform the evaluation using \nickname coupled with Alike-l~\cite{Zhao2022ALIKE}.
The `first frame' initialization  choice is equivalent to adding increased noise to the starting guess, with increasing error as the target pose moves far away from the initial pose (at the first frame). 
However, this approach does ensure reproducibility and provides a consistent baseline for fair comparisons with future work, offering a reliable measure of our method’s robustness.
Our results indicate that after three iterations of the \pnpransac paradigm, our method converges to a low localization error, even when starting from less accurate poses than those provided by DenseVLAD~\cite{torii2015densevlad}.

    \begin{table*}[t]
\begin{center}
        \begin{tabular}{l|c|c|c|c|c|c}
        \toprule
            Scene & Iteration & \makecell{Feature\\Extractor} & \makecell{Initial\\pose error\\$(cm, deg)$} &
            \makecell{Estimated\\pose error\\$(cm, deg)$} & \makecell{Avg. N. of\\inliers} & \makecell{Success\\rate (\%)} \\
            \midrule
chess & 1st & alike-l & 21.92, 12.13 & 0.77, 0.21 & 66 & 100.00 \\
             & 2nd &              &      -     & 0.74, 0.19 & 74 & 100.00 \\
             & 3rd &              &      -     & 0.72, 0.18 & 74 & 100.00 \\
\cline{2-7}
 & 1st & alike-n & 21.92, 12.13 & 0.73, 0.21 & 64 & 100.00 \\
             & 2nd &              &      -     & 0.68, 0.18 & 71 & 100.00 \\
             & 3rd &              &      -     & 0.64, 0.17 & 71 & 100.00 \\
\cline{2-7}
 & 1st & alike-s & 21.92, 12.13 & 0.82, 0.26 & 117 & 100.00 \\
             & 2nd &              &      -     & 0.76, 0.22 & 122 & 100.00 \\
             & 3rd &              &      -     & 0.71, 0.20 & 122 & 100.00 \\
\cline{2-7}
 & 1st & alike-t & 21.92, 12.13 & 1.01, 0.29 & 66 & 100.00 \\
             & 2nd &              &      -     & 0.94, 0.27 & 73 & 100.00 \\
             & 3rd &              &      -     & 0.89, 0.25 & 73 & 100.00 \\
\cline{2-7}
 & 1st & SuperPoint & 0.22, 12.13 & 0.68, 0.19 & 88 & 100.00 \\
             & 2nd &              &      -     & 0.67, 0.17 & 99 & 100.00 \\
             & 3rd &              &      -     & 0.64, 0.16 & 99 & 100.00 \\
\midrule
fire & 1st & alike-l & 34.37, 13.23 & 0.82, 0.34 & 73 & 100.00 \\
             & 2nd &              &      -     & 0.88, 0.36 & 72 & 100.00 \\
             & 3rd &              &      -     & 0.83, 0.34 & 72 & 99.85 \\
\cline{2-7}
 & 1st & alike-n & 34.37, 13.23 & 0.86, 0.37 & 66 & 100.00 \\
             & 2nd &              &      -     & 0.93, 0.37 & 66 & 100.00 \\
             & 3rd &              &      -     & 0.89, 0.35 & 66 & 99.60 \\
\cline{2-7}
 & 1st & alike-s & 34.37, 13.23 & 1.22, 0.47 & 172 & 100.00 \\
             & 2nd &              &      -     & 1.64, 0.60 & 162 & 100.00 \\
             & 3rd &              &      -     & 1.52, 0.56 & 163 & 99.70 \\
\cline{2-7}
 & 1st & alike-t & 34.37, 13.23 & 1.17, 0.47 & 59 & 100.00 \\
             & 2nd &              &      -     & 1.37, 0.52 & 55 & 100.00 \\
             & 3rd &              &      -     & 1.30, 0.49 & 55 & 99.55 \\
\cline{2-7}
 & 1st & SuperPoint & 34.37, 13.23 & 1.02, 0.39 & 67 & 100.00 \\
             & 2nd &              &      -     & 1.04, 0.38 & 67 & 100.00 \\
             & 3rd &              &      -     & 0.98, 0.36 & 67 & 98.65 \\
\midrule
heads & 1st & alike-l & 15.77, 14.97 & 0.73, 0.46 & 46 & 100.00 \\
             & 2nd &              &      -     & 0.67, 0.41 & 53 & 100.00 \\
             & 3rd &              &      -     & 0.66, 0.40 & 53 & 94.50 \\
\cline{2-7}
 & 1st & alike-n & 15.77, 14.97 & 1.08, 0.59 & 38 & 100.00 \\
             & 2nd &              &      -     & 0.97, 0.53 & 43 & 100.00 \\
             & 3rd &              &      -     & 0.96, 0.56 & 43 & 91.30 \\
\cline{2-7}
 & 1st & alike-s & 15.77, 14.97 & 0.70, 0.43 & 81 & 100.00 \\
             & 2nd &              &      -     & 0.62, 0.37 & 92 & 100.00 \\
             & 3rd &              &      -     & 0.59, 0.36 & 92 & 99.20 \\
\cline{2-7}
 & 1st & alike-t & 15.77, 14.97 & 0.89, 0.52 & 52 & 100.00 \\
             & 2nd &              &      -     & 0.81, 0.48 & 59 & 100.00 \\
             & 3rd &              &      -     & 0.76, 0.44 & 59 & 98.90 \\
\cline{2-7}
 & 1st & SuperPoint & 15.77, 14.97 & 0.62, 0.39 & 76 & 100.00 \\
             & 2nd &              &      -     & 0.54, 0.34 & 87 & 100.00 \\
             & 3rd &              &      -     & 0.52, 0.32 & 88 & 99.20 \\
\midrule
\multicolumn{7}{c}{Continue on next page} \\
\bottomrule
        \end{tabular}
\end{center}
\end{table*}
\begin{table*}
\begin{center}
        \begin{tabular}{l|c|c|c|c|c|c}
\midrule
office & 1st & alike-l & 28.58, 11.06 & 1.69, 0.43 & 36 & 100.00 \\
             & 2nd &              &      -     & 1.68, 0.41 & 39 & 100.00 \\
             & 3rd &              &      -     & 1.63, 0.39 & 40 & 99.25 \\
\cline{2-7}
 & 1st & alike-n & 28.58, 11.06 & 1.77, 0.47 & 35 & 100.00 \\
             & 2nd &              &      -     & 1.74, 0.45 & 37 & 100.00 \\
             & 3rd &              &      -     & 1.69, 0.42 & 37 & 97.78 \\
\cline{2-7}
 & 1st & alike-s & 28.58, 11.06 & 1.63, 0.45 & 69 & 100.00 \\
             & 2nd &              &      -     & 1.56, 0.41 & 72 & 100.00 \\
             & 3rd &              &      -     & 1.55, 0.40 & 72 & 99.98 \\
\cline{2-7}
 & 1st & alike-t & 28.58, 11.06 & 2.57, 0.68 & 37 & 100.00 \\
             & 2nd &              &      -     & 2.26, 0.61 & 42 & 100.00 \\
             & 3rd &              &      -     & 2.21, 0.58 & 42 & 99.15 \\
\cline{2-7}
 & 1st & SuperPoint & 28.58, 11.06 & 1.75, 0.43 & 65 & 100.00 \\
             & 2nd &              &      -     & 1.71, 0.41 & 72 & 100.00 \\
             & 3rd &              &      -     & 1.64, 0.37 & 72 & 99.75 \\
\midrule
pumpkin & 1st & alike-l & 31.38, 10.81 & 1.39, 0.30 & 69 & 100.00 \\
             & 2nd &              &      -     & 1.39, 0.29 & 76 & 100.00 \\
             & 3rd &              &      -     & 1.31, 0.28 & 76 & 98.65 \\
\cline{2-7}
 & 1st & alike-n & 31.38, 10.81 & 1.58, 0.36 & 70 & 100.00 \\
             & 2nd &              &      -     & 1.53, 0.34 & 76 & 100.00 \\
             & 3rd &              &      -     & 1.46, 0.31 & 76 & 93.45 \\
\cline{2-7}
 & 1st & alike-s & 31.38, 10.81 & 1.38, 0.31 & 118 & 100.00 \\
             & 2nd &              &      -     & 1.38, 0.29 & 121 & 100.00 \\
             & 3rd &              &      -     & 1.34, 0.28 & 122 & 99.25 \\
\cline{2-7}
 & 1st & alike-t & 31.38, 10.81 & 1.87, 0.43 & 81 & 100.00 \\
             & 2nd &              &      -     & 1.70, 0.39 & 90 & 100.00 \\
             & 3rd &              &      -     & 1.67, 0.37 & 91 & 96.40 \\
\cline{2-7}
 & 1st & SuperPoint & 31.38, 10.81 & 1.50, 0.33 & 110 & 100.00 \\
             & 2nd &              &      -     & 1.51, 0.31 & 120 & 100.00 \\
             & 3rd &              &      -     & 1.45, 0.29 & 120 & 99.05 \\
\midrule
redkitchen & 1st & alike-l & 29.38, 11.97 & 1.23, 0.30 & 45 & 100.00 \\
             & 2nd &              &      -     & 1.18, 0.25 & 54 & 100.00 \\
             & 3rd &              &      -     & 1.15, 0.24 & 54 & 98.08 \\
\cline{2-7}
 & 1st & alike-n & 29.38, 11.97 & 1.37, 0.33 & 51 & 100.00 \\
             & 2nd &              &      -     & 1.34, 0.32 & 60 & 100.00 \\
             & 3rd &              &      -     & 1.21, 0.28 & 60 & 96.02 \\
\cline{2-7}
 & 1st & alike-s & 29.38, 11.97 & 4.66, 1.09 & 57 & 100.00 \\
             & 2nd &              &      -     & 4.28, 1.00 & 67 & 100.00 \\
             & 3rd &              &      -     & 4.03, 0.94 & 68 & 77.42 \\
\cline{2-7}
 & 1st & alike-t & 29.38, 11.97 & 1.44, 0.31 & 57 & 100.00 \\
             & 2nd &              &      -     & 1.39, 0.29 & 66 & 100.00 \\
             & 3rd &              &      -     & 1.33, 0.27 & 67 & 99.38 \\
\cline{2-7}
 & 1st & SuperPoint & 0.29, 11.97 & 1.38, 0.30 & 79 & 100.00 \\
             & 2nd &              &      -     & 1.42, 0.27 & 93 & 100.00 \\
             & 3rd &              &      -     & 1.33, 0.24 & 93 & 98.74 \\
\midrule
\multicolumn{7}{c}{Continue on next page} \\
\bottomrule
        \end{tabular}
\end{center}
\end{table*}
\begin{table*}
\begin{center}
        \begin{tabular}{l|c|c|c|c|c|c}
\midrule
stairs & 1st & alike-l & 26.19, 15.81 & 3.80, 1.03 & 11 & 100.00 \\
             & 2nd &              &      -     & 3.02, 0.81 & 12 & 100.00 \\
             & 3rd &              &      -     & 2.74, 0.82 & 12 & 97.90 \\
\cline{2-7}
 & 1st & alike-n & 26.19, 15.81 & 7.19, 1.97 & 10 & 100.00 \\
             & 2nd &              &      -     & 6.28, 1.65 & 10 & 100.00 \\
             & 3rd &              &      -     & 5.96, 1.59 & 10 & 93.10 \\
\cline{2-7}
 & 1st & alike-s & 26.19, 15.81 & 5.78, 1.63 & 70 & 100.00 \\
             & 2nd &              &      -     & 5.18, 1.49 & 68 & 100.00 \\
             & 3rd &              &      -     & 5.03, 1.51 & 68 & 100.00 \\
\cline{2-7}
 & 1st & alike-t & 26.19, 15.81 & 5.30, 1.49 & 14 & 100.00 \\
             & 2nd &              &      -     & 4.38, 1.20 & 15 & 100.00 \\
             & 3rd &              &      -     & 4.03, 1.07 & 15 & 100.00 \\
\cline{2-7}
 & 1st & SuperPoint & 26.19, 15.81 & 5.83, 1.69 & 27 & 100.00 \\
             & 2nd &              &      -     & 4.54, 1.21 & 31 & 100.00 \\
             & 3rd &              &      -     & 4.05, 1.07 & 31 & 99.90 \\
\midrule
Overall & 1st & alike-t & 26.80, 12.85 & 2.03, 0.60 & 52 & 99.89 \\
Average & 2nd &              &      -     & 1.83, 0.54 & 57 & 99.24 \\
             & 3rd &              &      -     & 1.74, 0.50 & 57 & 99.05 \\
\cline{2-7}
             & 1st & alike-s & 26.80, 12.85 & 2.31, 0.66 & 97 & 97.83 \\
             & 2nd &              &      -     & 2.20, 0.63 & 100 & 96.83 \\
             & 3rd &              &      -     & 2.11, 0.61 & 101 & 96.51 \\
\cline{2-7}
             & 1st & alike-n & 26.80, 12.85 & 2.08, 0.62 & 47 & 97.18 \\
             & 2nd &              &      -     & 1.93, 0.55 & 51 & 96.28 \\
             & 3rd &              &      -     & 1.83, 0.52 & 51 & 95.89 \\
\cline{2-7}
             & 1st & alike-l & 26.80, 12.85 & 1.49, 0.44 & 49 & 99.25 \\
             & 2nd &              &      -     & 1.37, 0.39 & 54 & 98.65 \\
             & 3rd &              &      -     & 1.29, 0.38 & 54 & 98.32 \\
\cline{2-7}
             & 1st & SuperPoint & 26.80, 12.85 & 1.82, 0.53 & 73 & 99.81 \\
             & 2nd &              &      -     & 1.63, 0.44 & 81 & 99.49 \\
             & 3rd &              &      -     & 1.52, 0.40 & 81 & 99.33 \\
 \bottomrule
 \end{tabular}
    \end{center}
    \caption{
    \textbf{6-DoF median localization errors on the 7-Scenes dataset~\cite{shotton13cvpr}} for the various features extractors used to train \nickname.}
    \label{tab:svfr_comparison_7scenes}
\end{table*}

    \begin{table*}[t]
\begin{center}
        \setlength{\tabcolsep}{4pt}
\begin{tabular}[t]{l|c|c|c|c|c|c}
            \hline
            Scene & Iteration & \makecell{Feature\\Extractor} & \makecell{Initial\\pose error\\$(cm, deg)$} &
            \makecell{Estimated\\pose error\\$(cm, deg)$} & \makecell{Avg. N. of\\inliers}& \makecell{Success\\rate (\%)} \\
            \midrule
Shop         & 1st & alike-l & 136.31, 7.19 & 5.38, 0.27 & 208 & 100.00 \\
             & 2nd &              &      -     & 5.63, 0.22 & 231 & 100.00 \\
             & 3rd &              &      -     & 5.48, 0.25 & 231 & 100.00 \\
\cline{2-7}
             & 1st & alike-n & 136.31, 7.19 & 5.27, 0.28 & 185 & 100.00 \\
             & 2nd &              &      -     & 5.43, 0.23 & 208 & 100.00 \\
             & 3rd &              &      -     & 5.09, 0.24 & 208 & 100.00 \\
\cline{2-7}
             & 1st & alike-s & 136.31, 7.19 & 5.99, 0.24 & 225 & 100.00 \\
             & 2nd &              &      -     & 5.76, 0.25 & 250 & 100.00 \\
             & 3rd &              &      -     & 6.05, 0.25 & 250 & 100.00 \\
\cline{2-7}
             & 1st & alike-t & 136.31, 7.19 & 5.65, 0.27 & 203 & 100.00 \\
             & 2nd &              &      -     & 5.89, 0.26 & 224 & 100.00 \\
             & 3rd &              &      -     & 5.25, 0.25 & 224 & 100.00 \\
\cline{2-7}
             & 1st & SuperPoint & 136.31, 7.19 & 5.87, 0.29 & 204 & 100.00 \\
             & 2nd &              &      -     & 5.20, 0.26 & 225 & 100.00 \\
             & 3rd &              &      -     & 5.47, 0.26 & 224 & 100.00 \\
\midrule
College      & 1st & alike-l & 289.98, 5.96 & 18.19, 0.25 & 359 & 100.00 \\
             & 2nd &              &      -     & 17.04, 0.26 & 373 & 100.00 \\
             & 3rd &              &      -     & 15.25, 0.23 & 372 & 100.00 \\
\cline{2-7}
             & 1st & alike-n & 289.98, 5.96 & 16.82, 0.28 & 315 & 100.00 \\
             & 2nd &              &      -     & 17.38, 0.28 & 327 & 100.00 \\
             & 3rd &              &      -     & 17.61, 0.26 & 327 & 100.00 \\
\cline{2-7}
             & 1st & alike-s & 289.98, 5.96 & 16.64, 0.27 & 327 & 100.00 \\
             & 2nd &              &      -     & 15.74, 0.26 & 338 & 100.00 \\
             & 3rd &              &      -     & 15.67, 0.24 & 338 & 100.00 \\
\cline{2-7}
             & 1st & alike-t & 289.98, 5.96 & 17.64, 0.28 & 326 & 100.00 \\
             & 2nd &              &      -     & 16.33, 0.26 & 336 & 100.00 \\
             & 3rd &              &      -     & 16.52, 0.25 & 337 & 100.00 \\
\cline{2-7}
             & 1st & superpoint & 289.98, 5.96 & 17.88, 0.27 & 326 & 100.00 \\
             & 2nd &              &      -     & 18.15, 0.28 & 336 & 100.00 \\
             & 3rd &              &      -     & 17.52, 0.27 & 336 & 100.00 \\

\midrule
Great        & 1st & alike-l & 719.21, 9.47 & 32.46, 0.16 & 103 & 100.00 \\
             & 2nd &              &      -     & 29.48, 0.15 & 116 & 100.00 \\
             & 3rd &              &      -     & 27.40, 0.14 & 116 & 99.87 \\
\cline{2-7}
             & 1st & alike-n & 719.21, 9.47 & 37.88, 0.21 & 82 & 100.00 \\
             & 2nd &              &      -     & 35.20, 0.18 & 91 & 100.00 \\
             & 3rd &              &      -     & 32.05, 0.18 & 91 & 99.21 \\
\cline{2-7}
             & 1st & alike-s & 719.21, 9.47 & 36.18, 0.19 & 94 & 100.00 \\
             & 2nd &              &      -     & 34.27, 0.18 & 105 & 100.00 \\
             & 3rd &              &      -     & 31.78, 0.16 & 106 & 99.87 \\
\cline{2-7}
             & 1st & alike-t & 719.21, 9.47 & 33.78, 0.19 & 101 & 100.00 \\
             & 2nd &              &      -     & 31.33, 0.15 & 114 & 100.00 \\
             & 3rd &              &      -     & 28.83, 0.14 & 114 & 100.00 \\
\cline{2-7}
             & 1st & SuperPoint & 719.21, 9.47 & 34.69, 0.22 & 142 & 100.00 \\
             & 2nd &              &      -     & 30.71, 0.20 & 161 & 100.00 \\
             & 3rd &              &      -     & 29.09, 0.20 & 161 & 100.00 \\
\midrule
\multicolumn{7}{c}{Continue on next page} \\
\bottomrule
        \end{tabular}
\end{center}
\end{table*}
\begin{table*}
\begin{center}
        \begin{tabular}{l|c|c|c|c|c|c}
\midrule
Hospital     & 1st & alike-l & 405.22, 7.58 & 22.18, 0.44 & 155 & 100.00 \\
             & 2nd &              &      -     & 21.37, 0.40 & 160 & 100.00 \\
             & 3rd &              &      -     & 19.37, 0.36 & 160 & 100.00 \\
\cline{2-7}
             & 1st & alike-n & 405.22, 7.58 & 27.28, 0.47 & 128 & 100.00 \\
             & 2nd &              &      -     & 22.68, 0.44 & 132 & 100.00 \\
             & 3rd &              &      -     & 21.17, 0.40 & 131 & 100.00 \\
\cline{2-7}
             & 1st & alike-s & 405.22, 7.58 & 25.13, 0.44 & 132 & 100.00 \\
             & 2nd &              &      -     & 25.71, 0.47 & 136 & 100.00 \\
             & 3rd &              &      -     & 20.75, 0.37 & 136 & 100.00 \\
\cline{2-7}
             & 1st & alike-t & 405.22, 7.58 & 26.30, 0.51 & 140 & 100.00 \\
             & 2nd &              &      -     & 25.10, 0.48 & 145 & 100.00 \\
             & 3rd &              &      -     & 20.14, 0.41 & 145 & 100.00 \\
\cline{2-7}
             & 1st & SuperPoint & 405.22, 7.58 & 31.53, 0.56 & 143 & 100.00 \\
             & 2nd &              &      -     & 31.05, 0.55 & 148 & 100.00 \\
             & 3rd &              &      -     & 27.46, 0.54 & 148 & 100.00 \\
\midrule
Church       & 1st & alike-l & 287.61, 9.36 & 11.58, 0.38 & 201 & 100.00 \\
             & 2nd &              &      -     & 10.31, 0.31 & 228 & 100.00 \\
             & 3rd &              &      -     & 10.35, 0.30 & 228 & 100.00 \\
\cline{2-7}
             & 1st & alike-n & 287.61, 9.36 & 12.46, 0.43 & 181 & 100.00 \\
             & 2nd &              &      -     & 11.53, 0.35 & 206 & 100.00 \\
             & 3rd &              &      -     & 10.90, 0.33 & 206 & 100.00 \\
\cline{2-7}
             & 1st & alike-s & 287.61, 9.36 & 12.67, 0.42 & 180 & 100.00 \\
             & 2nd &              &      -     & 12.01, 0.36 & 203 & 100.00 \\
             & 3rd &              &      -     & 11.40, 0.35 & 204 & 99.81 \\
\cline{2-7}
             & 1st & alike-t & 287.61, 9.36 & 12.18, 0.40 & 170 & 100.00 \\
             & 2nd &              &      -     & 11.66, 0.35 & 192 & 100.00 \\
             & 3rd &              &      -     & 11.21, 0.36 & 192 & 100.00 \\
\cline{2-7}
             & 1st & SuperPoint & 287.61, 9.36 & 14.15, 0.49 & 188 & 100.00 \\
             & 2nd &              &      -     & 12.72, 0.42 & 220 & 100.00 \\
             & 3rd &              &      -     & 11.43, 0.38 & 220 & 99.81 \\
\midrule
Overall      & 1st & alike-t & 367.67, 7.91 & 19.11, 0.33 & 188 & 100.00 \\
Average      & 2nd &              &      -     & 18.06, 0.30 & 202 & 100.00 \\
             & 3rd &              &      -     & 16.39, 0.28 & 202 & 100.00 \\
\cline{2-7}
             & 1st & alike-s & 367.67, 7.91 & 19.32, 0.31 & 192 & 100.00 \\
             & 2nd &              &      -     & 18.70, 0.30 & 206 & 100.00 \\
             & 3rd &              &      -     & 17.13, 0.27 & 207 & 99.94 \\
\cline{2-7}
             & 1st & alike-n & 367.67, 7.91 & 19.94, 0.34 & 178 & 100.00 \\
             & 2nd &              &      -     & 18.44, 0.30 & 193 & 99.92 \\
             & 3rd &              &      -     & 17.36, 0.28 & 193 & 99.84 \\
\cline{2-7}
             & 1st & alike-l & 367.67, 7.91 & 17.96, 0.30 & 205 & 100.00 \\
             & 2nd &              &      -     & 16.77, 0.27 & 222 & 100.00 \\
             & 3rd &              &      -     & 15.57, 0.26 & 221 & 99.97 \\
\cline{2-7}
             & 1st & SuperPoint & 367.67, 7.91 & 20.82, 0.37 & 201 & 100.00 \\
             & 2nd &              &      -     & 19.57, 0.34 & 218 & 99.96 \\
             & 3rd &              &      -     & 18.19, 0.33 & 218 & 99.96 \\
\bottomrule
\end{tabular}
    \end{center}
    \caption{\textbf{6-DoF median localization errors on the Cambridge dataset~\cite{Kendall15iccv}} for the various features extractors used to train \nickname.}
    \label{tab:svfr_comparison_cambridge}
\end{table*}


\end{document}